\documentclass{article}

    \PassOptionsToPackage{numbers, compress}{natbib}

  \usepackage[main, final]{neurips_2026}

\usepackage[utf8]{inputenc} %
\usepackage[T1]{fontenc}    %
\usepackage{hyperref}       %
\usepackage{url}            %
\usepackage{booktabs}       %
\usepackage{amsfonts}       %
\usepackage{nicefrac}       %
\usepackage{microtype}      %
\usepackage{xcolor}         %
\definecolor{ImpUp}{HTML}{0066CC}
\definecolor{ImpDown}{HTML}{CC0000}
\usepackage{graphicx}
\usepackage{capt-of}
\usepackage{amsmath}
\usepackage{cleveref}
\usepackage{wrapfig}
\usepackage{caption}
\usepackage[table]{xcolor}
\usepackage{multirow}
\usepackage{subcaption}
\usepackage{listings}
\usepackage[breakable]{tcolorbox}
\usepackage{enumitem}
\usepackage{makecell}

\title{Reading the Unreadable: Text-Aware Image Super-Resolution Needs Reasoning}

\author{%
  Jaeseong Lee \quad
  Jinwoo Kim \quad 
  Jinho Jeong \quad 
  Seon Joo Kim \\[4pt]
  CIPLAB, Yonsei University\\[6pt]
  {\normalfont\small\ttfamily Project Page:\ \href{https://jasonleex1995.github.io/TAISR_needs_Reasoning}{https://jasonleex1995.github.io/TAISR\_needs\_Reasoning} } \\
}

\begin{document}

\maketitle

\vspace{-1.2em}
\begin{center}
    \includegraphics[width=0.97\linewidth]{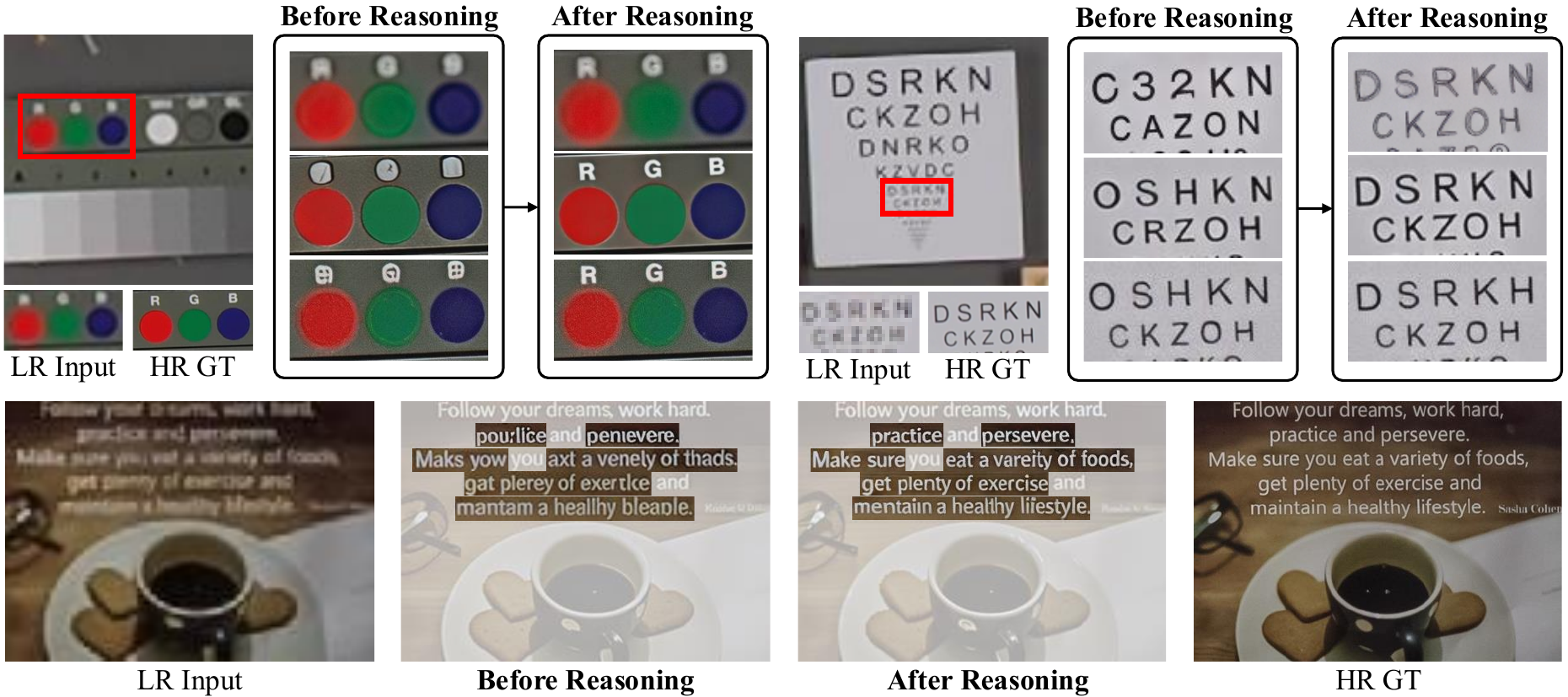}
\end{center}
\vspace{-0.6em}
\begingroup
\captionsetup{font=small}
\captionof{figure}{\textbf{TAISR needs reasoning.} Each panel shows the LR input, baseline SR outputs (Before Reasoning), outputs with RTC (After Reasoning), and the HR ground truth. The highlighted text is unreadable from local pixels alone, so the baselines hallucinate characters. With reasoning, the same models recover most of it.}
\label{fig:teaser}
\endgroup
\vspace{1em}

\begin{abstract}

Text-aware image super-resolution (TAISR) aims to recover high-resolution images from low-resolution inputs while preserving textual content. Existing methods treat this as a visual restoration problem, yet severely degraded text often cannot be read from local visual evidence alone. Humans instead read it by reasoning over image context, logical patterns, and prior knowledge. This raises a fundamental question: \emph{Does TAISR need reasoning?} To answer this, we introduce two diagnostic tools. The first, ReasonText, is a human-curated benchmark that separates reasoning-required text from locally recoverable text. The second, Ground-Truth Text-Crop Accuracy (GTTCA), is a recognition-based metric that measures text restoration quality without text detection errors. Evaluating twelve recent super-resolution (SR) models, we find a consistent and substantial performance drop on reasoning-required text, revealing a systematic blind spot. Two observations point to a remedy. One is that text-conditioned SR models restore text substantially better when given ground-truth text. The other is that multimodal large language models (MLLMs) can read degraded text in low-resolution images through reasoning. We therefore propose Reasoning Transfer via Captioning (RTC), a training-free plug-in that uses an external MLLM to infer the degraded text and feeds the resulting caption to a text-conditioned SR model. RTC consistently improves all six text-conditioned SR models we evaluate, especially on reasoning-required text. Surprisingly, a real-world SR model with RTC can even outperform the TAISR model with RTC, suggesting that reasoning should be a core design axis for future TAISR.

\end{abstract}

\section{Introduction}
\label{sec:intro}

Real-world image super-resolution (real-world SR)~\cite{swinir, realesrgan} aims to recover high-resolution (HR) images from low-resolution (LR) inputs corrupted by unknown degradations. Recent advances~\cite{faithdiff, dit4sr, diffbir, stablesr, pure, seesr, supir} leverage text-to-image generation models~\cite{sd35, luminamgpt, sdxl, ldm} as generative backbones, producing visually plausible details even from severely degraded inputs. However, such generative priors often fail to preserve the structural information present in the LR image~\cite{faithdiff, dit4sr}. This issue is particularly severe for text, where altering a single stroke can change its meaning entirely. To address this challenge, a line of work has introduced text-aware image super-resolution (TAISR)~\cite{tadisr, unit, terediff, textsr}, which restores full scene images while preserving textual content. Unlike scene text image super-resolution (STISR)~\cite{stt, textzoom}, which operates on cropped text patches, TAISR inherits the open-scene setting of real-world SR and targets text fidelity within full images.

Existing TAISR methods augment the SR model with text-aware modules such as text detection~\cite{textsr}, segmentation~\cite{tadisr}, or spotting~\cite{unit, terediff}. Although these modules improve the localization and restoration of textual regions, they remain centered on local visual evidence. By contrast, humans read what appears unreadable. When a distant sign is nearly illegible, we can often still read it by reasoning from the surrounding context, logical patterns, or prior knowledge of brands, places, and vocabulary. Current TAISR pipelines lack explicit mechanisms for such reasoning and hallucinate characters where local evidence is insufficient (Figure~\ref{fig:teaser}). This gap raises the central question of this paper: \emph{Does TAISR need reasoning?} Answering it requires measurement, yet current tools fall short in two ways. Existing benchmarks~\cite{terediff} do not distinguish reasoning-required text from locally recoverable text, and their evaluation protocols rely on text spotting models~\cite{abcnetv2, testr}, whose detection errors contaminate the measurement of text restoration quality.

We address both shortcomings with two diagnostic tools. First, we introduce ReasonText, a human-curated benchmark that assigns each text instance a perception-grounded difficulty level, separating reasoning-required text from locally recoverable text. Second, we propose Ground-Truth Text-Crop Accuracy (GTTCA), a text restoration metric that runs text recognition directly on ground-truth text regions of SR outputs, bypassing the text detection stage. Using these tools, we evaluate twelve recent real-world SR and TAISR models and find that every model's accuracy drops sharply on reasoning-required text: the lowest GTTCA on locally recoverable text across all models (45.7\%) still exceeds the highest on reasoning-required text (24.5\%). This failure occurs regardless of architecture, generative backbone, or training paradigm, and persists even in models explicitly designed for text.

These results answer our question: \emph{TAISR does need reasoning.} What remains is how to supply it, and we start from two observations. First, text-conditioned SR models~\cite{faithdiff, dit4sr, unit, terediff, osediff, supir} restore reasoning-required text substantially better when the ground-truth transcripts are provided in their captions. Second, modern multimodal large language models (MLLMs)~\cite{gemma4, qwen35} can infer locally unreadable text in LR images through reasoning. Building on these observations, we propose Reasoning Transfer via Captioning (RTC). In RTC, an external MLLM infers the degraded text and writes it into a caption, which then replaces the default caption of a text-conditioned SR model. Since both models remain frozen, RTC is training-free and can be attached to any such model.

Empirically, RTC consistently improves both real-world SR and TAISR models on ReasonText, and these gains carry over to the existing TAISR benchmarks~\cite{terediff}, SA-Text-test and RealText. Surprisingly, the strongest real-world SR model~\cite{dit4sr} with RTC outperforms the strongest TAISR model~\cite{unit} with RTC, even though the latter builds on the same backbone~\cite{sd35} and adds text-aware modules. This result suggests that, once reasoning is supplied, current text-aware modules may bring no additional benefit. Reasoning should therefore be a core design axis for TAISR, and a natural next step is to build it into the text-aware modules themselves.

Our contributions are summarized as follows:
\begin{itemize}[itemsep=2pt, topsep=2pt]
    \item We introduce ReasonText, the first TAISR benchmark for evaluating reasoning-required text, and GTTCA, a metric that measures text restoration quality free of text detection errors.
    \item We evaluate twelve recent real-world SR and TAISR models on ReasonText and show that every model's accuracy drops sharply on reasoning-required text.
    \item We propose RTC, a training-free plug-in that injects an external MLLM's reasoning into text-conditioned SR models through captions. Extensive experiments show that RTC consistently improves both real-world SR and TAISR models.
\end{itemize}

\section{Related Work}
\label{sec:related_work}

\noindent\textbf{Real-World Image Super-Resolution (Real-World SR).} Real-world SR aims to recover HR images from LR inputs corrupted by diverse, unknown degradations. Early GAN-based methods~\cite{swinir, realesrgan} improve perceptual quality with adversarial training~\cite{gan}, but often introduce visual artifacts. To avoid such artifacts and restore realistic details even under severe degradations, recent work instead builds on generative models with strong priors, including diffusion and autoregressive (AR) models. Multi-step diffusion-based methods~\cite{faithdiff, dit4sr, diffbir, stablesr, seesr, supir} adapt pretrained text-to-image diffusion models~\cite{sd35, sdxl, ldm} to real-world SR with conditioning modules such as ControlNet-style adapters~\cite{controlnet}. Few-step diffusion-based methods~\cite{sinsr, osediff, resshift} reduce the inference cost of iterative denoising by starting the diffusion process from the LR image instead of Gaussian noise, or by distilling a multi-step model into a single step. AR-based methods~\cite{realsrr1, pure} adapt a multimodal AR model~\cite{luminamgpt} with post-training techniques from large language models. However, such strong priors can also hallucinate plausible yet incorrect details. This failure mode is especially problematic for text, where even a minor deviation can alter the meaning.

\noindent\textbf{Text-Aware Image Super-Resolution (TAISR).} Motivated by this failure mode, TAISR aims to restore full scene images while explicitly preserving textual content. TextSR~\cite{textsr} first detects text regions with an OCR model, then restores the text with its own STISR model and the rest of the image with a real-world SR model~\cite{realesrgan}. TADiSR~\cite{tadisr} adds a text segmentation decoder to predict text masks jointly with the restored image. TeReDiff~\cite{terediff} and UniT~\cite{unit} use a text spotting module that takes diffusion features as input and predicts text to guide restoration. Despite these diverse designs, all of them remain centered on local visual evidence and do not consider the reasoning that humans use to read locally unreadable text.

\noindent\textbf{Multimodal Large Language Models (MLLMs).} Recent MLLMs~\cite{qwen3vl, gemma4, kimi25, qwen35} have demonstrated strong reasoning capabilities on benchmarks requiring joint visual and textual understanding~\cite{mathvista, mmmu}. In image super-resolution, however, MLLMs have been used in a much more limited way. Real-world SR methods~\cite{faithdiff, dit4sr, supir} employ LLaVA-1.5~\cite{llava15} to caption LR images, and UniT adopts Qwen2.5-VL~\cite{qwen25vl} to initialize text predictions. In both cases, the MLLM is an earlier-generation model with limited reasoning ability and serves only as a perceptual front-end. In contrast, we leverage the reasoning capability of recent MLLMs to infer locally unreadable text.

\section{TAISR Needs Reasoning}
\label{sec:diagnosis}

In this section, we examine whether existing real-world SR and TAISR models can restore text that requires reasoning. In Section~\ref{subsec:benchmark}, we introduce ReasonText, a benchmark that separates reasoning-required text from locally recoverable text. In Section~\ref{subsec:metric}, we propose Ground-Truth Text-Crop Accuracy (GTTCA), a metric that measures text restoration quality by recognizing text directly at ground-truth locations. Using these two tools, we evaluate twelve recent SR models in Section~\ref{subsec:failure} and find that all of them suffer a substantial accuracy drop on reasoning-required text.

\subsection{ReasonText Benchmark}
\label{subsec:benchmark}

\begin{figure*}[t]
    \centering
    \includegraphics[width=\linewidth]{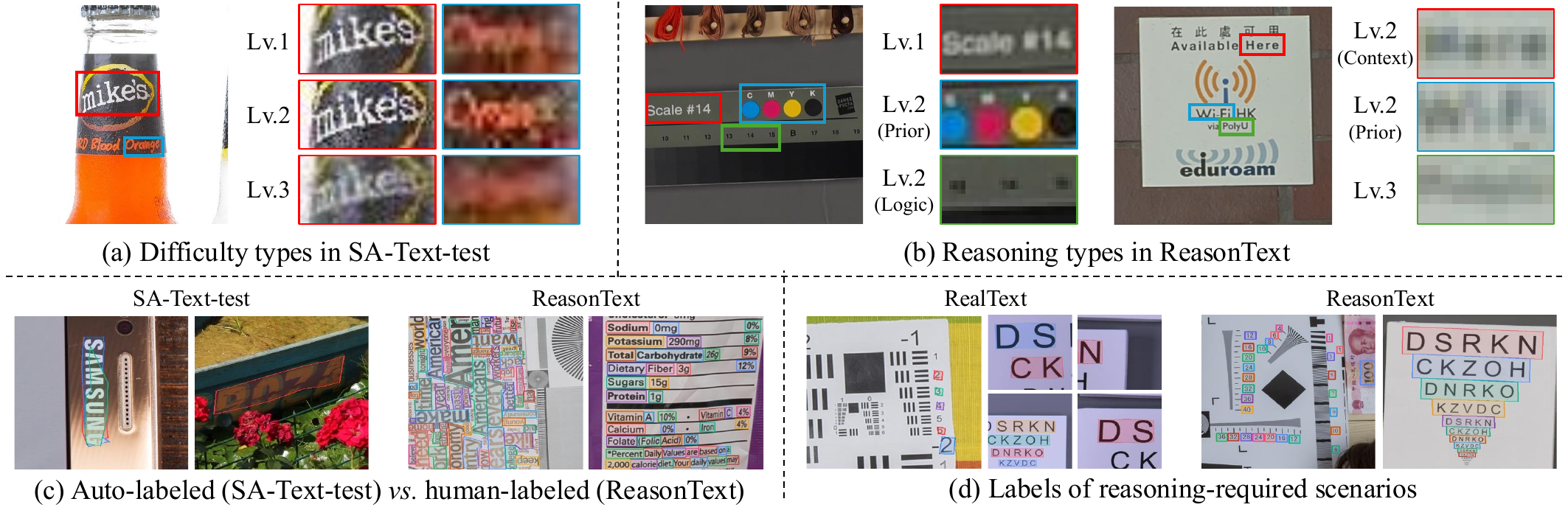}
    
    \caption{\textbf{ReasonText addresses gaps in existing TAISR benchmarks.} (a) SA-Text-test assigns difficulty at the image level via three degradation strengths, which does not indicate whether each instance requires reasoning. (b) ReasonText instead labels each instance with a perception-grounded difficulty level, separating reasoning-required from locally recoverable text. (c) ReasonText is constructed from human annotations, which remain reliable across diverse scenes. (d) ReasonText crops the image to preserve context needed to read reasoning-required text.}
    \label{fig:dataset}
    \vspace{-3pt} 
\end{figure*}

\paragraph{Limitations of existing benchmarks.}
To evaluate whether SR models can restore reasoning-required text, we need a \emph{reasoning-required label} for each LR text instance, but no existing benchmark provides one. SA-Text-test~\cite{terediff} might seem to meet this need at first glance, since it provides three LR versions of each HR image at different synthetic degradation strengths. However, degradation strength is an \emph{image-level} attribute, so it does not indicate whether an individual text instance requires reasoning (Figure~\ref{fig:dataset}(a)). A natural alternative is to add such labels to the existing benchmarks~\cite{terediff}, SA-Text-test and RealText, but two of their characteristics rule this out. First, their text polygons and transcripts contain non-trivial errors, as they are produced by fully automatic pipelines without human verification (Figure~\ref{fig:dataset}(c)). Second, their images are cropped without considering the context of each text instance, leaving few instances that can be read through reasoning (Figure~\ref{fig:dataset}(d)). We therefore construct ReasonText with three key properties: (1) all text polygons and transcripts are annotated by humans, (2) each image is cropped to preserve the context of its text instances, and (3) each instance is labeled with a \emph{perception-grounded} difficulty level: readable from the text region alone, readable only with reasoning, or unreadable (Figure~\ref{fig:dataset}(b)).

\paragraph{Data collection and annotation.}
We build \emph{ReasonText} on two real-world SR benchmarks, RealSR-v3~\cite{realsrv3} and DRealSR~\cite{drealsr}, which provide camera-captured LR--HR pairs rather than synthetically degraded ones. Trained annotators first label word-level polygons, transcripts, and orientations for every alphanumeric text instance in each full HR image, and the labels go through multiple rounds of verification. They then crop a square HR patch around each text instance, including any content related to that instance. Following the standard $\times 4$ SR setting~\cite{unit, terediff}, we resize each HR patch to $512\!\times\!512$ and its aligned LR patch to $128\!\times\!128$.

\paragraph{Difficulty labeling.}
We obtain the perception-grounded difficulty level of each text instance through a two-stage labeling process. Both stages follow the same protocol. Five of ten annotators are randomly assigned to each text instance, and transcribe the text only when they can read it, skipping it otherwise to prevent guessing. We then compare the collected transcripts with the ground-truth (GT) transcript, and regard a text instance as identified when they match.

\hangindent=1em\hangafter=0
\textbf{Stage 1 (cropped LR text).} We crop each text instance from its LR patch using the annotated polygon, yielding a \emph{cropped LR text image} that contains no context outside the text itself. Instances identified at this stage are assigned \textbf{Level 1}. Since they are read from the text alone, we regard Level~1 instances as \emph{locally recoverable text}.

\hangindent=1em\hangafter=0
\textbf{Stage 2 (full LR patch).} Stage~2 begins after Stage~1 is completed for all instances. We show each remaining instance again within its full LR patch, with its text region highlighted, so that annotators can use the surrounding context. When they can read the text, annotators also select which cue helped them most: \emph{image context}, when the cue comes from what else is visible in the patch, such as the same text appearing in another region; \emph{logical patterns}, when it comes from a regularity that the text follows, such as the numeric sequence on a ruler; or \emph{prior knowledge}, when it comes from what the annotator already knows, such as a brand logo. Instances identified at this stage are assigned \textbf{Level 2}. Since they are read only once the surrounding context is available, we regard Level~2 instances as \emph{reasoning-required text}. The rest are assigned \textbf{Level 3}. Since none of our annotators can read them even with the context, we treat Level~3 instances as \emph{unrecoverable text}.

For Level~3 instances, an SR output that happens to match the GT transcript cannot be distinguished from a lucky guess. We therefore exclude Level~3 instances from evaluation and discard LR--HR pairs containing only Level~3 instances. The resulting benchmark contains $513$ LR--HR pairs with $2{,}463$ text instances (Level~1: $1{,}754$, Level~2: $709$). Further details are provided in Appendix~\ref{sec:supp_benchmark}.

\subsection{GTTCA Metric}
\label{subsec:metric}
Existing TAISR works~\cite{unit, terediff} evaluate text restoration quality using text spotting models~\cite{abcnetv2, testr}, which jointly perform text detection and recognition on the SR output. Detection errors therefore contaminate the measurement of text restoration quality, since an instance missed by the detector counts as a failure even if its text is well restored. Building on the reliable annotations of ReasonText, we propose \emph{Ground-Truth Text-Crop Accuracy} (GTTCA), which extends the recognition-based evaluation of STISR~\cite{stt, textzoom} to the full-image setting of TAISR. GTTCA crops each text instance with its ground-truth (GT) polygon, removing the detection stage from the measurement.

GTTCA is computed in four steps (see Figure~\ref{fig:supp_fig_metric} in Appendix). For each text instance, we (1)~crop the SR output around its GT polygon with padding, (2)~rotate and flip the crop according to its GT orientation so that the text reads upright, (3)~recognize the text in the crop with an OCR model, and (4)~check whether the recognized text exactly matches the GT transcript after normalization, following the standard STISR protocol~\cite{stt, textzoom}. GTTCA is the fraction of matched instances. We report GTTCA separately on Level~1 and Level~2 instances, as well as over both levels (Overall), using PaddleOCR-VL-1.5~\cite{paddleocrvl15} as the OCR model. Further details are provided in Appendix~\ref{sec:supp_metric}.

\begin{table}[t]
\centering
\definecolor{restgray}{gray}{0.5}
\setlength{\aboverulesep}{0pt}
\setlength{\belowrulesep}{0pt}
\renewcommand{\arraystretch}{1.25}
\caption{\textbf{Quantitative results of twelve real-world SR and TAISR models on ReasonText.} Every model struggles to recover Level~2 (reasoning-required) text. For each metric, the top-three models are in black (\textbf{best}, \underline{second-best}, third-best plain) and the rest in \textcolor{restgray}{gray}. The bottom rows report Pearson/Spearman correlations between each metric and overall GTTCA across the twelve SR models. $\uparrow$ ($\downarrow$): higher (lower) is better.}
\label{tab:tab_failure}
\resizebox{\textwidth}{!}{%
\begin{tabular}{cl|cc|cc|ccc|ccc}
\toprule
\multirow{3}{*}{Group} & \multirow{3}{*}{Model} & \multicolumn{2}{c|}{\multirow{2}{*}{Fidelity}} & \multicolumn{5}{c|}{Perceptual} & \multicolumn{3}{c}{Text Restoration} \\
\cmidrule(lr){5-9} \cmidrule(lr){10-12}
& & \multicolumn{2}{c|}{} & \multicolumn{2}{c|}{Reference} & \multicolumn{3}{c|}{No Reference} & \multicolumn{3}{c}{GTTCA} \\
\cmidrule(lr){3-4} \cmidrule(lr){5-6} \cmidrule(lr){7-9} \cmidrule(lr){10-12}
& & PSNR-Y$\uparrow$ & SSIM-Y$\uparrow$ & LPIPS$\downarrow$ & DISTS$\downarrow$ & MANIQA$\uparrow$ & MUSIQ$\uparrow$ & CLIP-IQA$\uparrow$ & Overall$\uparrow$ & Level~1$\uparrow$ & Level~2$\uparrow$ \\
\midrule
\rowcolor{cyan!20}
\multicolumn{2}{c|}{HR (oracle)} & 80.0000 & 1.0000 & 0.0000 & 0.0000 & 0.4261 & 66.2367 & 0.5403 & 94.6813 & 96.2942 & 90.6911 \\
\midrule
\multirow{2}{*}{A} & Real-ESRGAN & \textcolor{restgray}{24.0453} & \underline{0.7694} & \textcolor{restgray}{0.2134} & \textcolor{restgray}{0.1719} & \textcolor{restgray}{0.4463} & \textcolor{restgray}{68.5397} & \textcolor{restgray}{0.4611} & \textcolor{restgray}{40.3167} & \textcolor{restgray}{52.9647} & \textcolor{restgray}{9.0268} \\
 & SwinIR & 24.7762 & \textbf{0.7859} & \textbf{0.1989} & \underline{0.1626} & \textcolor{restgray}{0.4288} & \textcolor{restgray}{66.9483} & \textcolor{restgray}{0.4357} & \textcolor{restgray}{45.6760} & \textcolor{restgray}{59.4071} & \textcolor{restgray}{11.7066} \\
\midrule
\multirow{4}{*}{B} & DiffBIR & \textcolor{restgray}{24.1097} & \textcolor{restgray}{0.7039} & \textcolor{restgray}{0.2591} & \textcolor{restgray}{0.1880} & \textbf{0.6127} & \textbf{71.8427} & \underline{0.6417} & \textcolor{restgray}{54.9736} & 69.6123 & \textcolor{restgray}{18.7588} \\
 & SUPIR & \textcolor{restgray}{24.2971} & \textcolor{restgray}{0.7098} & \textcolor{restgray}{0.2622} & \textcolor{restgray}{0.1947} & \textcolor{restgray}{0.4608} & \textcolor{restgray}{63.5998} & \textcolor{restgray}{0.5922} & 55.2984 & \textcolor{restgray}{68.5291} & 22.5670 \\
 & FaithDiff & \textcolor{restgray}{24.2038} & \textcolor{restgray}{0.7363} & \underline{0.2061} & \textbf{0.1623} & \textcolor{restgray}{0.5095} & \underline{71.6540} & \textcolor{restgray}{0.5821} & \textcolor{restgray}{52.0097} & \textcolor{restgray}{65.1653} & \textcolor{restgray}{19.4640} \\
 & DiT4SR & \textcolor{restgray}{22.6539} & \textcolor{restgray}{0.6933} & \textcolor{restgray}{0.2477} & \textcolor{restgray}{0.1830} & \textcolor{restgray}{0.4973} & \textcolor{restgray}{70.8502} & \textcolor{restgray}{0.6074} & \underline{59.1555} & \underline{73.1471} & \textbf{24.5416} \\
\midrule
\multirow{3}{*}{C} & ResShift & \underline{24.9170} & \textcolor{restgray}{0.7645} & \textcolor{restgray}{0.2219} & \textcolor{restgray}{0.1839} & \textcolor{restgray}{0.4110} & \textcolor{restgray}{63.3033} & \textcolor{restgray}{0.4902} & \textcolor{restgray}{47.7060} & \textcolor{restgray}{61.0034} & \textcolor{restgray}{14.8096} \\
 & SinSR & \textbf{25.1644} & 0.7668 & 0.2116 & \textcolor{restgray}{0.1747} & \textcolor{restgray}{0.4551} & \textcolor{restgray}{66.1994} & \textcolor{restgray}{0.5574} & \textcolor{restgray}{43.1181} & \textcolor{restgray}{55.9863} & \textcolor{restgray}{11.2835} \\
 & OSEDiff & \textcolor{restgray}{23.9596} & \textcolor{restgray}{0.7533} & \textcolor{restgray}{0.2147} & 0.1666 & \textcolor{restgray}{0.4803} & 71.1559 & \textcolor{restgray}{0.5672} & \textcolor{restgray}{44.6204} & \textcolor{restgray}{58.4379} & \textcolor{restgray}{10.4372} \\
\midrule
D & PURE & \textcolor{restgray}{22.2970} & \textcolor{restgray}{0.6706} & \textcolor{restgray}{0.2652} & \textcolor{restgray}{0.1893} & 0.5233 & \textcolor{restgray}{70.0036} & \textbf{0.6531} & \textcolor{restgray}{34.3890} & \textcolor{restgray}{45.7241} & \textcolor{restgray}{6.3470} \\
\midrule
\multirow{2}{*}{E} & TeReDiff & \textcolor{restgray}{23.5188} & \textcolor{restgray}{0.7233} & \textcolor{restgray}{0.2362} & \textcolor{restgray}{0.1796} & \underline{0.5638} & \textcolor{restgray}{70.5181} & \textcolor{restgray}{0.6183} & \textcolor{restgray}{51.2383} & \textcolor{restgray}{65.1083} & \textcolor{restgray}{16.9252} \\
 & UniT & \textcolor{restgray}{22.7672} & \textcolor{restgray}{0.6503} & \textcolor{restgray}{0.2814} & \textcolor{restgray}{0.2072} & \textcolor{restgray}{0.5159} & \textcolor{restgray}{68.7382} & 0.6315 & \textbf{61.3886} & \textbf{76.4538} & \underline{24.1185} \\
\midrule
\multicolumn{2}{c|}{Pearson $r$ vs. Overall} & $-0.1566$ & $-0.4414$ & $+0.4333$ & $+0.4459$ & $+0.3175$ & $+0.0806$ & $+0.3528$ & \multicolumn{3}{c}{--} \\
\multicolumn{2}{c|}{Spearman $\rho$ vs. Overall} & $-0.1329$ & $-0.5455$ & $+0.4196$ & $+0.4056$ & $+0.3147$ & $+0.1958$ & $+0.3636$ & \multicolumn{3}{c}{--} \\
\bottomrule
\end{tabular}%
}
\end{table}

\subsection{SR Models Fail on Reasoning-Required Text}
\label{subsec:failure}
We evaluate twelve recent publicly available real-world SR and TAISR models on ReasonText, covering five groups: (A)~GAN-based~\cite{swinir, realesrgan}, (B)~multi-step diffusion-based~\cite{faithdiff, dit4sr, diffbir, supir}, (C)~few-step diffusion-based~\cite{sinsr, osediff, resshift}, (D)~AR-based~\cite{pure}, and (E)~TAISR\footnote{Some prior TAISR models such as TextSR~\cite{textsr} and TADiSR~\cite{tadisr} are not publicly released and are therefore excluded from our comparison.}~\cite{unit, terediff}. We report fidelity (PSNR and SSIM on the Y channel), perceptual quality (reference-based LPIPS~\cite{lpips} and DISTS~\cite{dists}, and no-reference MANIQA~\cite{maniqa}, MUSIQ~\cite{musiq}, and CLIP-IQA~\cite{clipiqa}), and text restoration quality (GTTCA). We also report all metrics on the HR images as a reference (HR oracle). Table~\ref{tab:tab_failure} reports the results, and qualitative results are shown in Appendix~\ref{sec:supp_failure}.

\textbf{Finding 1: All models struggle on reasoning-required text.} GTTCA drops sharply from Level~1 to Level~2 for all twelve models, regardless of architecture, training paradigm, or whether the model is explicitly text-aware. The drop is large enough that the score ranges of the two levels do not overlap. The lowest Level~1 score ($45.72\%$) still exceeds the highest Level~2 score ($24.54\%$). Since the HR oracle reaches $90.69\%$ on Level~2, far above every SR model, the gap reflects a limitation of the SR models rather than of the text recognizer behind the metric.

\textbf{Finding 2: Current text-aware design helps only on locally recoverable text.} The TAISR model UniT~\cite{unit} achieves the best overall GTTCA ($61.39\%$), ahead of the strongest real-world SR model, DiT4SR~\cite{dit4sr} ($59.16\%$). Since the two build on the same backbone~\cite{sd35}, this comparison largely isolates the effect of UniT's text-aware design. The margin, however, comes entirely from Level~1, where UniT leads by over three points ($76.45\%$ versus $73.15\%$), while on Level~2 it vanishes ($24.12\%$ versus $24.54\%$). This matches our earlier argument that text-aware modules remain centered on local visual evidence, so their benefit does not extend to reasoning-required text.

\textbf{Finding 3: Better conventional SR quality does not mean better text restoration.} We compute the Pearson and Spearman rank correlation between each conventional metric and overall GTTCA across the twelve models. None of them tracks text restoration quality. PSNR and SSIM show no positive correlation with GTTCA. LPIPS and DISTS, for which lower is better, correlate in the wrong direction. No-reference image quality assessment (NR-IQA) metrics correlate only weakly. Improving conventional SR quality therefore does not carry over to text restoration, which has to be measured on its own.

\noindent Together, these findings answer the question we posed: \emph{TAISR does need reasoning.} What remains is how to supply it, which we address next.

\section{Reasoning Transfer via Captioning}
\label{sec:method}

\begin{table}[t]
\centering
\setlength{\aboverulesep}{0pt}
\setlength{\belowrulesep}{0pt}
\renewcommand{\arraystretch}{1.25}
\caption{\textbf{Injecting ground-truth text improves GTTCA for all six text-conditioned SR models.} Each model is evaluated with its original caption and with the GT transcripts of its Level 1 and Level 2 instances inserted as a natural-language caption. In the + GT rows, \textcolor{ImpUp}{blue}/\textcolor{ImpDown}{red} indicate an improvement/decline over the original caption. $\uparrow$($\downarrow$): higher (lower) is better.}
\label{tab:tab_gt}
\resizebox{\textwidth}{!}{%
\begin{tabular}{cl|cc|cc|ccc|ccc}
\toprule
\multirow{3}{*}{Group} & \multirow{3}{*}{Model} & \multicolumn{2}{c|}{\multirow{2}{*}{Fidelity}} & \multicolumn{5}{c|}{Perceptual} & \multicolumn{3}{c}{Text Restoration} \\
\cmidrule(lr){5-9} \cmidrule(lr){10-12}
& & \multicolumn{2}{c|}{} & \multicolumn{2}{c|}{Reference} & \multicolumn{3}{c|}{No Reference} & \multicolumn{3}{c}{GTTCA} \\
\cmidrule(lr){3-4} \cmidrule(lr){5-6} \cmidrule(lr){7-9} \cmidrule(lr){10-12}
& & PSNR-Y$\uparrow$ & SSIM-Y$\uparrow$ & LPIPS$\downarrow$ & DISTS$\downarrow$ & MANIQA$\uparrow$ & MUSIQ$\uparrow$ & CLIP-IQA$\uparrow$ & Overall$\uparrow$ & Level~1$\uparrow$ & Level~2$\uparrow$ \\
\midrule
\rowcolor{cyan!20}
\multicolumn{2}{c|}{HR (oracle)} & 80.0000 & 1.0000 & 0.0000 & 0.0000 & 0.4261 & 66.2367 & 0.5403 & 94.6813 & 96.2942 & 90.6911 \\
\midrule
\multirow{6}{*}{B} & SUPIR & 24.2971 & 0.7098 & 0.2622 & 0.1947 & 0.4608 & 63.5998 & 0.5922 & 55.2984 & 68.5291 & 22.5670 \\
 & \quad + GT & \textcolor{ImpUp}{24.5912} & \textcolor{ImpDown}{0.7038} & \textcolor{ImpDown}{0.2732} & \textcolor{ImpDown}{0.1964} & \textcolor{ImpDown}{0.4349} & \textcolor{ImpDown}{60.9064} & \textcolor{ImpDown}{0.5714} & \textcolor{ImpUp}{69.7686} & \textcolor{ImpUp}{81.6534} & \textcolor{ImpUp}{40.3667} \\
\cmidrule(l){2-12}
 & FaithDiff & 24.2038 & 0.7363 & 0.2061 & 0.1623 & 0.5095 & 71.6540 & 0.5821 & 52.0097 & 65.1653 & 19.4640 \\
 & \quad + GT & \textcolor{ImpUp}{24.4677} & \textcolor{ImpUp}{0.7467} & \textcolor{ImpUp}{0.1962} & \textcolor{ImpUp}{0.1567} & \textcolor{ImpDown}{0.4951} & \textcolor{ImpDown}{71.4650} & \textcolor{ImpDown}{0.5739} & \textcolor{ImpUp}{62.4523} & \textcolor{ImpUp}{75.3250} & \textcolor{ImpUp}{30.6065} \\
\cmidrule(l){2-12}
 & DiT4SR & 22.6539 & 0.6933 & 0.2477 & 0.1830 & 0.4973 & 70.8502 & 0.6074 & 59.1555 & 73.1471 & 24.5416 \\
 & \quad + GT & \textcolor{ImpUp}{23.0546} & \textcolor{ImpUp}{0.7182} & \textcolor{ImpUp}{0.2317} & \textcolor{ImpUp}{0.1797} & \textcolor{ImpDown}{0.4944} & \textcolor{ImpDown}{70.8355} & \textcolor{ImpDown}{0.5932} & \textcolor{ImpUp}{74.1048} & \textcolor{ImpUp}{85.1767} & \textcolor{ImpUp}{46.7137} \\
\midrule
\multirow{2}{*}{C} & OSEDiff & 23.9596 & 0.7533 & 0.2147 & 0.1666 & 0.4803 & 71.1559 & 0.5672 & 44.6204 & 58.4379 & 10.4372 \\
 & \quad + GT & \textcolor{ImpDown}{23.9394} & \textcolor{ImpUp}{0.7546} & \textcolor{ImpDown}{0.2179} & \textcolor{ImpDown}{0.1695} & \textcolor{ImpUp}{0.4811} & \textcolor{ImpDown}{71.1110} & \textcolor{ImpUp}{0.5676} & \textcolor{ImpUp}{51.4657} & \textcolor{ImpUp}{66.4082} & \textcolor{ImpUp}{14.4993} \\
\midrule
\multirow{4}{*}{E} & TeReDiff & 23.5188 & 0.7233 & 0.2362 & 0.1796 & 0.5638 & 70.5181 & 0.6183 & 51.2383 & 65.1083 & 16.9252 \\
 & \quad + GT & \textcolor{ImpDown}{23.1785} & \textcolor{ImpDown}{0.7176} & \textcolor{ImpDown}{0.2388} & \textcolor{ImpDown}{0.1812} & \textcolor{ImpUp}{0.5694} & \textcolor{ImpUp}{71.0449} & \textcolor{ImpUp}{0.6251} & \textcolor{ImpUp}{59.5128} & \textcolor{ImpUp}{72.7708} & \textcolor{ImpUp}{26.7137} \\
\cmidrule(l){2-12}
 & UniT & 22.7672 & 0.6503 & 0.2814 & 0.2072 & 0.5159 & 68.7382 & 0.6315 & 61.3886 & 76.4538 & 24.1185 \\
 & \quad + GT & \textcolor{ImpUp}{23.1021} & \textcolor{ImpUp}{0.6725} & \textcolor{ImpUp}{0.2694} & \textcolor{ImpUp}{0.1994} & \textcolor{ImpDown}{0.5024} & \textcolor{ImpDown}{68.3612} & \textcolor{ImpDown}{0.6169} & \textcolor{ImpUp}{70.6618} & \textcolor{ImpUp}{82.8848} & \textcolor{ImpUp}{40.4231} \\
\bottomrule
\end{tabular}%
}
\end{table}

Reasoning can happen inside the SR model or outside it. The internal approach gives the SR model itself the ability to reason. The external approach obtains reasoning from an external model and injects it into the SR model. The internal approach is appealing but currently hard to pursue. Finding~3 shows that the conventional quality criteria that SR training pursues do not track text restoration, and Finding~2 shows that even text-specific modules and objectives do not help on reasoning-required text. This leaves no obvious training signal for teaching an SR model to infer locally illegible text, which motivates our external approach.

This approach raises two questions.

\begin{itemize}[itemsep=2pt, topsep=2pt]
    \item[(i)] \textbf{Reasoning generation.} What kind of reasoning-aware representation should we extract from the LR image?
    \item[(ii)] \textbf{Injection.} How should this representation be injected into the SR model so that it restores reasoning-required text?
\end{itemize}

The design space for both questions is broad. A reasoning-aware representation could take many forms, such as a similar image retrieved from an external source, a text string inferred from the LR image, or features taken from the reasoning model itself. Each of these could in turn be injected in different ways, for example by concatenating it with the SR model's intermediate features, attending to it in a cross-attention layer, or passing it as the model's text input. Rather than exploring this space exhaustively, we derive a method from two empirical observations. The first settles both design questions, and the second makes the resulting choice practical.

\subsection{Observation 1: Ground-Truth Transcripts Improve Text Restoration}
\label{sec:injection}

Many recent SR models~\cite{faithdiff, dit4sr, unit, terediff, osediff, supir} already accept a text input, originally intended to convey high-level semantics or, in TAISR models, predicted text. Since this channel is already in place, we examine whether it can carry reasoning-aware information.

To test this, we inject the GT transcripts of ReasonText directly into the caption. Given the $n$ transcripts $t_1, t_2, \ldots, t_n$ of the Level 1 and Level 2 instances in an image, we replace the default caption with a natural-language caption listing them, of the form \texttt{"The image with the text \textquotedblleft $t_1$\textquotedblright, \textquotedblleft $t_2$\textquotedblright, $\ldots$, \textquotedblleft $t_n$\textquotedblright."}, and re-evaluate on ReasonText. Table~\ref{tab:tab_gt} shows that GTTCA improves for every model on both levels. The gains on Level~2 are substantial. DiT4SR~\cite{dit4sr} gains over $22$ points, from $24.54\%$ to $46.71\%$, and SUPIR~\cite{supir} and UniT~\cite{unit} gain over $16$ points each. As a side effect, NR-IQA scores drop for several models. We attribute the drop to the caption no longer describing the scene, since these models are trained with scene-describing captions. Adding a scene description largely restores NR-IQA (Appendix~\ref{sec:supp_rtc_injection}), and the RTC caption therefore includes one alongside the inferred text (Section~\ref{sec:rtc}).

This observation settles both design questions. For injection, we pass the reasoning-aware representation through the text channel that text-conditioned SR models already provide. For reasoning generation, the reasoning-aware representation is a natural-language caption listing the GT transcripts of the image. In a real-world scenario, however, the GT transcripts of an LR image are unavailable, so we need a way to approximate this caption from the LR image alone, which we address next.

\begin{table}[t]
\centering
\setlength{\aboverulesep}{0pt}
\setlength{\belowrulesep}{0pt}
\renewcommand{\arraystretch}{1.2}
\caption{\textbf{The reasoning prompt substantially improves MLLMs on reasoning-required (Level 2) text, where OCR models and the no-reasoning prompt lag far behind.} (Left) Text recall of dedicated OCR models. (Right) Text recall of MLLMs under the no-reasoning / reasoning prompts. Best per column in \textbf{bold}. $\uparrow$: higher is better.}

\label{tab:tab_ocr_mllm}
\begin{minipage}[t]{0.40\textwidth}
\centering
\resizebox{\textwidth}{!}{%
\begin{tabular}{l|ccc}
\toprule
Model & Overall$\uparrow$ & Level~1$\uparrow$ & Level~2$\uparrow$ \\
\midrule
PaddleOCR-VL-1.5 & 48.03 & 54.90 & 31.03 \\
TESTR & 56.44 & 61.29 & \textbf{44.43} \\
GLM-OCR & \textbf{59.32} & \textbf{66.08} & 42.60 \\
\bottomrule
\end{tabular}%
}
\end{minipage}
\hfill
\begin{minipage}[t]{0.58\textwidth}
\centering
\resizebox{\textwidth}{!}{%
\begin{tabular}{l|ccc|ccc}
\toprule
\multirow{2}{*}{Model} & \multicolumn{3}{c|}{no-reasoning} & \multicolumn{3}{c}{reasoning} \\
\cmidrule(lr){2-4} \cmidrule(lr){5-7}
& Overall$\uparrow$ & Level~1$\uparrow$ & Level~2$\uparrow$ & Overall$\uparrow$ & Level~1$\uparrow$ & Level~2$\uparrow$ \\
\midrule
Qwen2.5-VL 7B & 46.98 & 56.84 & 22.57 & 58.51 & 66.08 & 39.77 \\
Qwen3.5 27B & 58.42 & 69.67 & 30.61 & 77.51 & 83.81 & 61.92 \\
Gemma 4 31B & \textbf{69.18} & \textbf{78.62} & \textbf{45.84} & \textbf{82.01} & \textbf{86.72} & \textbf{70.38} \\
\bottomrule
\end{tabular}%
}
\end{minipage}
\end{table}

\subsection{Observation 2: MLLMs Can Approximate Ground-Truth Transcripts}
\label{sec:reasoning}

\paragraph{Choosing a reasoning generator.}

How can we obtain GT transcripts from an LR image? We measure a model's ability to do so with text recall, the fraction of GT transcripts that exactly match a word in its output, after both are normalized. We start with OCR models~\cite{paddleocrvl15, glmocr, testr}, the standard tool for reading text from images, and run them on the LR images of ReasonText. Table~\ref{tab:tab_ocr_mllm} shows that their recall drops sharply on Level~2, where even the best model reaches only $44.43\%$. Despite the rich priors from large image-text corpora, OCR models struggle with reasoning-required text, which makes them unsuitable as a reasoning generator.

We therefore turn to MLLMs, which read text as OCR models do but can also reason about the image~\cite{mathvista, mmmu}. We run Qwen2.5-VL~\cite{qwen25vl}, which UniT adopts, together with two more recent MLLMs, Qwen3.5~\cite{qwen35} and Gemma~4~\cite{gemma4}. Each model is run under two prompts. The \emph{no-reasoning prompt} instructs the model to list the text it can see, and the \emph{reasoning prompt} additionally instructs it to infer text using image context, logical patterns, or prior knowledge.

Table~\ref{tab:tab_ocr_mllm} shows that the reasoning prompt improves every model at both levels, and the gain is larger on Level~2 than on Level~1 in all three cases. Gemma~4 reaches the highest score, $70.38\%$ on Level~2, which exceeds the best OCR model by $25.95$ points. It misses some reasoning-required text, but reads a large portion of it. We therefore adopt Gemma~4 with the reasoning prompt as our reasoning generator, and use the transcripts it reads to build the caption.

\paragraph{Does the gain come from reasoning?}
One might ask whether the high Level~2 recall of Gemma~4 comes from something other than reasoning, and two explanations stand out. It may be exceptional local text recognition, strong enough to read from the text region alone what our annotators could not. Or it may be memorization, since ReasonText is built on the publicly available RealSR-v3 and DRealSR, which the model could have seen during pre-training.

To test the first, we mask everything outside the GT text regions, which removes the surrounding scene while leaving the text pixels intact, and run the same model on the masked images. The effect differs sharply between the two levels. Level~1 recall drops only slightly, from $86.72\%$ to $82.55\%$, since locally recoverable text does not need the surroundings. In contrast, Level~2 recall collapses, from $70.38\%$ to $29.90\%$. If Gemma~4 were simply a strong local text recognizer, masking the surroundings would barely change Level~2 recall, just as it barely changes Level~1 recall. The collapse rules out the first explanation, since Gemma~4 clearly needs the surroundings to read reasoning-required text.

\begin{figure*}[t]
    \centering
    \includegraphics[width=0.95\linewidth]{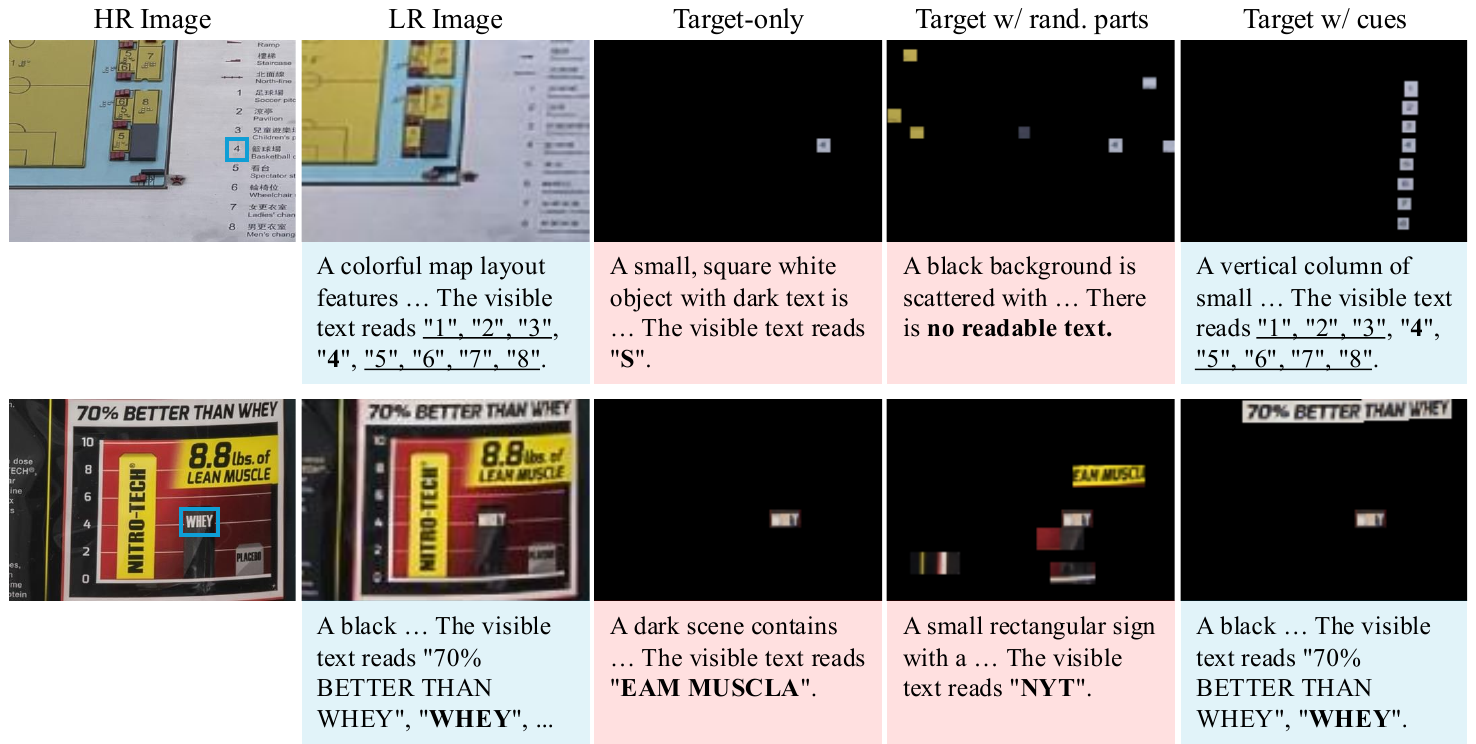}
    \caption{\textbf{Gemma 4 reads locally illegible text by reasoning over relevant cues.} Top: given only the target region (the locally illegible "4"), Gemma 4 fails. Adding random unrelated patches does not help, but adding the related numeric column "1"–"8" restores the correct reading. Bottom: likewise, "WHEY" is read only when the banner "70\% BETTER THAN WHEY" is visible. Blue and red backgrounds mark correct and incorrect readings.} 
    \label{fig:fig_mllm_mask}
\end{figure*}

To test the second, we vary the visible surrounding region (Figure~\ref{fig:fig_mllm_mask}). Gemma~4 recovers the correct transcript on \textbf{Target w/ cues}, where only the target text and the relevant cues are visible. In contrast, it fails on \textbf{Target w/ rand. parts}, where the cues are replaced by random regions of the same area, and on \textbf{Target-only}, the masked setting used above. This argues against the second explanation, since a model recalling a memorized image would not depend on the cue region in particular. With both alternatives unlikely, reasoning over the surroundings remains the most plausible explanation.

\subsection{Method: Reasoning Transfer via Captioning}
\label{sec:rtc}

Section~\ref{sec:injection} showed that a caption carrying the GT transcripts improves the restoration of reasoning-required text, and Section~\ref{sec:reasoning} showed that an MLLM with the reasoning prompt can infer most of those transcripts from the LR image. Putting the two together gives \emph{Reasoning Transfer via Captioning} (RTC). Given an LR image $\mathbf{x}_{\text{LR}}$, RTC proceeds in two steps.

\begin{enumerate}
    \item \textbf{Caption by reasoning.} Query an MLLM with $\mathbf{x}_{\text{LR}}$ and the reasoning prompt, which returns a two sentence caption $c$: a scene description followed by the inferred transcripts.
    \item \textbf{Restore by conditioning.} Pass $(\mathbf{x}_{\text{LR}}, c)$ to a text-conditioned SR model, with $c$ replacing its default caption, to produce the restored image $\hat{\mathbf{x}}_{\text{HR}}$.
\end{enumerate}

Since both models remain frozen and the caption $c$ is the only interface between them, RTC is training-free and attaches to any text-conditioned SR model, whether it was designed for real-world SR or for TAISR. This interface also makes failures traceable, since we can check whether the MLLM misread the text or the SR model failed to render it.

\section{Experiments}
\label{sec:experiments}

\subsection{Experimental Settings}
\label{sec:main_results_settings}

\begin{figure*}[t]
    \centering
    \includegraphics[width=0.95\linewidth]{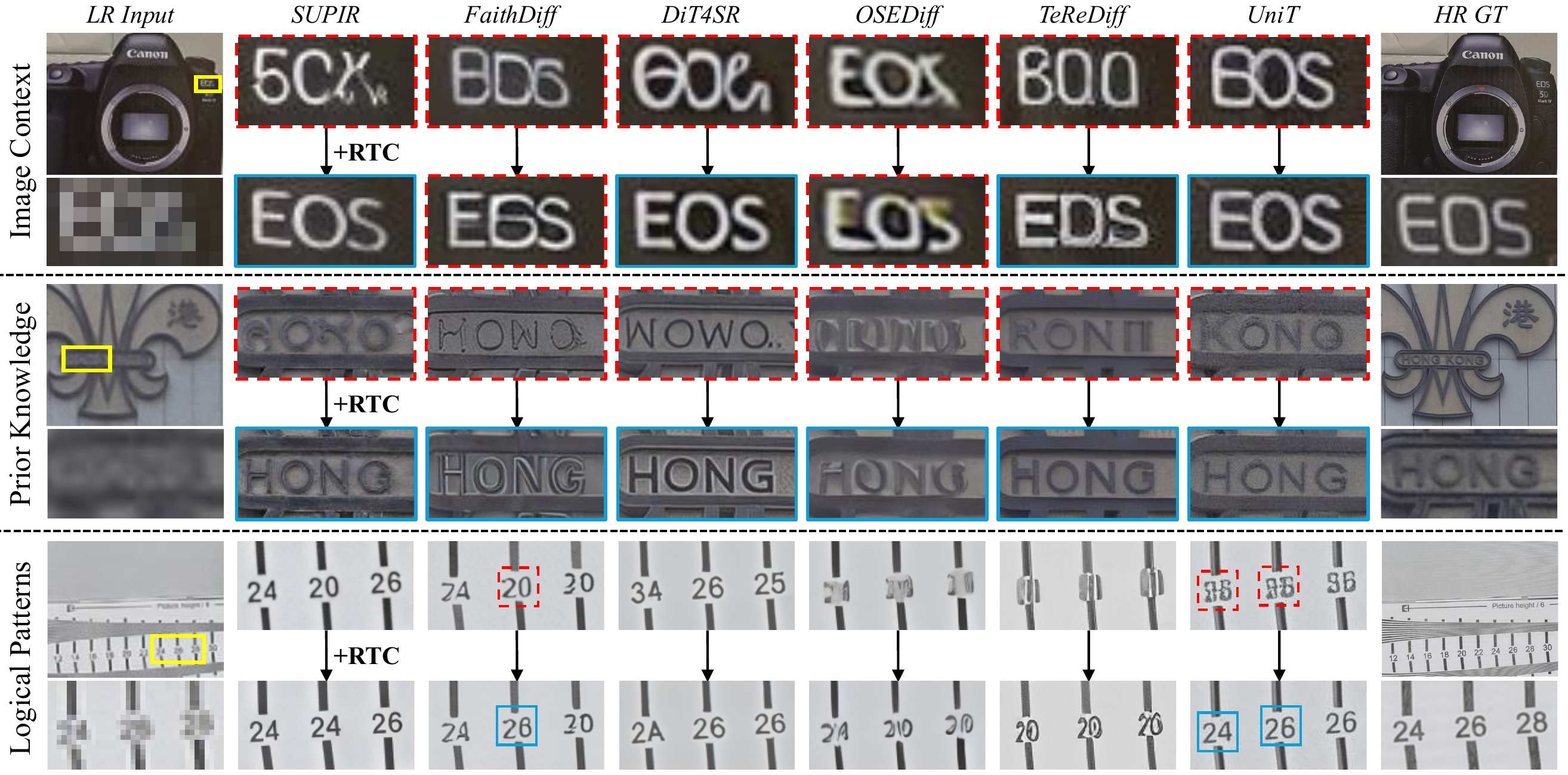}
    \caption{\textbf{Qualitative results of RTC across reasoning types.} For Level 2 instances of each reasoning type, the top and bottom rows show the outputs of six text-conditioned SR models without and with RTC, respectively. With RTC, some models can turn incorrect text into correct text.
    } 
    \label{fig:fig_rtc_sample}
\end{figure*}

We attach RTC to the six text-conditioned SR models of Section~\ref{sec:injection}. For each model, the baseline is its official inference pipeline with its default caption source, such as LLaVA-1.5 for DiT4SR and Qwen2.5-VL for UniT. \emph{+RTC} replaces only this caption with the one Gemma~4 produces under the reasoning prompt, leaving everything else unchanged. For a controlled comparison, we precompute one caption per LR image and feed the same caption to all six models, keeping it fixed throughout denoising. We evaluate on ReasonText with the metrics of Section~\ref{subsec:failure} and run all experiments on NVIDIA H200 GPUs. Table~\ref{tab:tab_rtc} reports the results, and Figure~\ref{fig:fig_rtc_sample} shows qualitative comparisons.

\begin{table}[t]
\centering
\setlength{\aboverulesep}{0pt}
\setlength{\belowrulesep}{0pt}
\renewcommand{\arraystretch}{1.25}
\caption{\textbf{Quantitative comparison of text-conditioned SR models with and without RTC on ReasonText.} RTC consistently improves text restoration for all six models. Baselines use each model's default caption source. In the + RTC rows, \textcolor{ImpUp}{blue}/\textcolor{ImpDown}{red} indicate an improvement/decline over the baseline. $\uparrow$($\downarrow$): higher (lower) is better.}
\label{tab:tab_rtc}
\resizebox{\textwidth}{!}{%
\begin{tabular}{cl|cc|cc|ccc|ccc}
\toprule
\multirow{3}{*}{Group} & \multirow{3}{*}{Model} & \multicolumn{2}{c|}{\multirow{2}{*}{Fidelity}} & \multicolumn{5}{c|}{Perceptual} & \multicolumn{3}{c}{Text Restoration} \\
\cmidrule(lr){5-9} \cmidrule(lr){10-12}
& & \multicolumn{2}{c|}{} & \multicolumn{2}{c|}{Reference} & \multicolumn{3}{c|}{No Reference} & \multicolumn{3}{c}{GTTCA} \\
\cmidrule(lr){3-4} \cmidrule(lr){5-6} \cmidrule(lr){7-9} \cmidrule(lr){10-12}
& & PSNR-Y$\uparrow$ & SSIM-Y$\uparrow$ & LPIPS$\downarrow$ & DISTS$\downarrow$ & MANIQA$\uparrow$ & MUSIQ$\uparrow$ & CLIP-IQA$\uparrow$ & Overall$\uparrow$ & Level~1$\uparrow$ & Level~2$\uparrow$ \\
\midrule
\rowcolor{cyan!20}
\multicolumn{2}{c|}{HR (oracle)} & 80.0000 & 1.0000 & 0.0000 & 0.0000 & 0.4261 & 66.2367 & 0.5403 & 94.6813 & 96.2942 & 90.6911 \\
\midrule
\multirow{6}{*}{B} & SUPIR & 24.2971 & 0.7098 & 0.2622 & 0.1947 & 0.4608 & 63.5998 & 0.5922 & 55.2984 & 68.5291 & 22.5670 \\
 & \quad + RTC & \textcolor{ImpDown}{23.9534} & \textcolor{ImpDown}{0.7071} & \textcolor{ImpUp}{0.2480} & \textcolor{ImpUp}{0.1846} & \textcolor{ImpUp}{0.5202} & \textcolor{ImpUp}{66.8668} & \textcolor{ImpUp}{0.6412} & \textcolor{ImpUp}{67.3163} & \textcolor{ImpUp}{78.8483} & \textcolor{ImpUp}{38.7870} \\
\cmidrule(l){2-12}
 & FaithDiff & 24.2038 & 0.7363 & 0.2061 & 0.1623 & 0.5095 & 71.6540 & 0.5821 & 52.0097 & 65.1653 & 19.4640 \\
 & \quad + RTC & \textcolor{ImpUp}{24.2724} & \textcolor{ImpUp}{0.7387} & \textcolor{ImpUp}{0.2020} & \textcolor{ImpUp}{0.1602} & \textcolor{ImpDown}{0.4975} & \textcolor{ImpDown}{71.4577} & \textcolor{ImpDown}{0.5797} & \textcolor{ImpUp}{61.0231} & \textcolor{ImpUp}{74.0023} & \textcolor{ImpUp}{28.9140} \\
\cmidrule(l){2-12}
 & DiT4SR & 22.6539 & 0.6933 & 0.2477 & 0.1830 & 0.4973 & 70.8502 & 0.6074 & 59.1555 & 73.1471 & 24.5416 \\
 & \quad + RTC & \textcolor{ImpUp}{22.6816} & \textcolor{ImpUp}{0.6936} & \textcolor{ImpDown}{0.2479} & 0.1830 & \textcolor{ImpUp}{0.5009} & \textcolor{ImpDown}{70.6745} & \textcolor{ImpUp}{0.6110} & \textcolor{ImpUp}{73.5688} & \textcolor{ImpUp}{84.7777} & \textcolor{ImpUp}{45.8392} \\
\midrule
\multirow{2}{*}{C} & OSEDiff & 23.9596 & 0.7533 & 0.2147 & 0.1666 & 0.4803 & 71.1559 & 0.5672 & 44.6204 & 58.4379 & 10.4372 \\
 & \quad + RTC & \textcolor{ImpDown}{23.8903} & \textcolor{ImpDown}{0.7504} & \textcolor{ImpDown}{0.2171} & \textcolor{ImpDown}{0.1682} & \textcolor{ImpUp}{0.4886} & \textcolor{ImpUp}{71.2281} & \textcolor{ImpUp}{0.5767} & \textcolor{ImpUp}{49.4113} & \textcolor{ImpUp}{64.2531} & \textcolor{ImpUp}{12.6939} \\
\midrule
\multirow{4}{*}{E} & TeReDiff & 23.5188 & 0.7233 & 0.2362 & 0.1796 & 0.5638 & 70.5181 & 0.6183 & 51.2383 & 65.1083 & 16.9252 \\
 & \quad + RTC & \textcolor{ImpDown}{23.3884} & \textcolor{ImpDown}{0.7212} & \textcolor{ImpDown}{0.2377} & \textcolor{ImpUp}{0.1795} & \textcolor{ImpUp}{0.5777} & \textcolor{ImpUp}{70.9852} & \textcolor{ImpUp}{0.6237} & \textcolor{ImpUp}{56.6382} & \textcolor{ImpUp}{70.1254} & \textcolor{ImpUp}{23.2722} \\
\cmidrule(l){2-12}
 & UniT & 22.7672 & 0.6503 & 0.2814 & 0.2072 & 0.5159 & 68.7382 & 0.6315 & 61.3886 & 76.4538 & 24.1185 \\
 & \quad + RTC & \textcolor{ImpUp}{22.8425} & \textcolor{ImpUp}{0.6555} & \textcolor{ImpUp}{0.2751} & \textcolor{ImpUp}{0.2041} & \textcolor{ImpUp}{0.5169} & \textcolor{ImpUp}{68.8130} & \textcolor{ImpUp}{0.6338} & \textcolor{ImpUp}{70.9704} & \textcolor{ImpUp}{82.5542} & \textcolor{ImpUp}{42.3131} \\
\bottomrule
\end{tabular}%
}
\end{table}

\subsection{Results}
\label{sec:main_results}

\paragraph{RTC recovers most of the gains observed with GT transcripts.}
RTC replaces the GT transcripts with transcripts inferred by an MLLM, so we first check whether the observations made with GT transcripts still hold under this replacement. Table~\ref{tab:tab_rtc} shows that they do. RTC improves GTTCA for every model on both levels, whether the model was designed for real-world SR or for TAISR. As with GT transcripts, the gains are larger on Level~2 than on Level~1 for most models, reaching $21.30$ points for DiT4SR (from $24.54\%$ to $45.84\%$). Moreover, the scene description in the RTC caption largely avoids the NR-IQA drop observed with transcript-only captions, and fidelity and reference-based perceptual metrics also remain stable. RTC is thus a practical method that needs neither GT transcripts nor training, and it improves the restoration of reasoning-required text, where all SR models struggle, while preserving general image quality.

\paragraph{With RTC, a real-world SR model surpasses a TAISR model.}
Without RTC, UniT achieves the best overall GTTCA, ahead of DiT4SR, which shares the same backbone but has no text-aware module. With RTC, however, the order reverses. DiT4SR\,+\,RTC outperforms UniT\,+\,RTC on both levels, even though both receive the same caption. This suggests that the current text-aware design is not complementary to reasoning, since a model without it performs better once reasoning is supplied. We therefore argue for rethinking text-aware design with reasoning as its core axis.

\paragraph{Generalization and robustness of RTC.}
The gains of RTC are not specific to ReasonText. On the existing TAISR benchmarks, SA-Text-test and RealText, RTC improves text restoration for all six models under their original evaluation protocol. RTC is also robust to imperfect captions. Even when Gemma~4 misreads the text with spelling errors, RTC still improves GTTCA, since the SR models can often restore the correct text. RTC therefore does not lower text restoration accuracy even when the MLLM misreads the text. Detailed results are provided in Appendix~\ref{sec:supp_exp}.

\section{Conclusion and Future Work}
\label{sec:conclusion}

We asked whether text-aware image super-resolution needs reasoning. To answer this, we introduced ReasonText, a benchmark that separates reasoning-required text from locally recoverable text, and GTTCA, a metric that measures text restoration quality without detection errors. With these tools, we found that twelve recent SR models, including those designed for text, all struggle on reasoning-required text. As a remedy, we proposed RTC, a training-free plug-in, based on two observations: (i) text-conditioned SR models restore text substantially better when their captions contain the GT transcripts, and (ii) MLLMs can infer these transcripts from the LR image through reasoning. In our experiments, RTC consistently improves text restoration for both real-world SR and TAISR models, especially on reasoning-required text. Moreover, with RTC, a real-world SR model surpasses a TAISR model built on the same backbone, which suggests that reasoning should be a core axis of text-aware design.

While our study shows that TAISR needs reasoning, several limitations remain. First, ReasonText is human-curated and built on existing real-world SR benchmarks, so it is limited in size and scope. It contains $513$ LR--HR pairs of natural images and annotates only alphanumeric text, leaving other scripts unannotated. Future work can extend it to more languages, a larger scale, more diverse domains, and revisit Level 3 with a larger or more specialized annotator pool. Second, even the best model with RTC remains far below the HR oracle on Level~2 ($45.84\%$ versus $90.69\%$). Most of this gap therefore lies not in reading the text but in rendering it, calling for SR models that faithfully render the given text as well as better ways of bringing reasoning into SR, including internal approaches. Finally, RTC applies only to text-conditioned SR models and cannot be used with SR models that have no text channel. A more general way of injecting reasoning, applicable to any SR model, is left for future work.

\section*{Acknowledgement}
\label{sec:acknowledgement}

This work was supported by Samsung Electronics Co., Ltd. (IO251211-14355-01), Samsung Research Funding
\& Incubation Center of Samsung Electronics under Project Number SRFC-IT2501-02, and Artificial Intelligence Graduate School Program, Yonsei University (No. RS-2020-II201361).

\bibliographystyle{plainnat}
\bibliography{refs}

\clearpage
\appendix

\setcounter{figure}{0}
\renewcommand{\thefigure}{\Alph{figure}}
\setcounter{table}{0}
\renewcommand{\thetable}{\Alph{table}}

\section{Additional Details on TAISR Needs Reasoning}

\subsection{ReasonText Benchmark}
\label{sec:supp_benchmark}

This section details the construction of ReasonText (Section~\ref{subsec:benchmark}).

\paragraph{HR image annotation.}
Three annotators participate in HR image annotation, and both annotation tasks below follow the same procedure. One annotator labels all instances in a pass, and a second annotator reviews the pass and corrects any label they find incorrect. The two then alternate, each reviewing the other's corrections, until a pass ends with no change. A third annotator reviews the converged labels, and any instance they flag is resolved by discussion among all three.

\begin{enumerate}
    \item \textbf{Text polygons.} Annotators draw a word-level polygon around every legible alphanumeric text instance in each HR image, including curved and perspective-distorted ones. Each polygon uses as many points as needed to enclose the text and nothing else.
    \item \textbf{Transcripts and orientations.} Annotators transcribe each text instance and record the transformation that brings it upright, consisting of a rotation from $\{0^\circ, 90^\circ, 180^\circ, 270^\circ\}$ and an optional horizontal flip.
\end{enumerate}

\paragraph{LR--HR patch construction.}
Existing TAISR benchmarks~\cite{terediff} obtain LR--HR pairs by random cropping. We find that random cropping often removes the surrounding context that humans use to read locally unreadable text, turning a reasoning-required instance into an unrecoverable one. We therefore crop patches manually, with three annotators participating in this stage.

For each text instance in an HR image, annotators draw a square crop that includes any content related to that instance, and record its coordinates. When no such content is present, the annotators draw a random square crop containing the instance. Two constraints apply: the crop must be at least $512\!\times\!512$, and it must enclose the target text polygon.

Since a crop is drawn for every text instance, many of the resulting crops overlap. Among overlapping crops with similar size, we automatically keep the one containing the most text instances and discard the others. We then map the coordinates of each retained crop to the $\times 4$ LR image provided by RealSR-v3~\cite{realsrv3} and DRealSR~\cite{drealsr} by dividing them by four, and crop the corresponding LR region. Since crop sizes vary, we finally resize the HR crop to $512\!\times\!512$ and the LR crop to $128\!\times\!128$, obtaining the standard $\times 4$ SR setting used in prior work~\cite{unit, terediff}.

\paragraph{HR patch re-annotation.}
Cropping can cut a text instance at the patch boundary, leaving a polygon that extends beyond the patch or a word that is only partially visible. We therefore re-annotate the instances that cross a patch boundary, adjusting each polygon to the visible extent and relabeling its transcript accordingly. This re-annotation follows the same procedure as the HR image annotation, with two annotators alternating corrections until a pass ends with no change, followed by a review by a third annotator whose flagged instances are resolved by discussion among all three.

\paragraph{Design choices for difficulty labeling.}
We make three core design choices that apply to both stages of Section~\ref{subsec:benchmark}. (1)~We assign a level to an instance when any of the five annotators identifies it, rather than requiring agreement among them. (2)~We select ten annotators for whom English is either a native language or a primary second language. Otherwise, an annotator unfamiliar with English would fail to read instances that others can read, pushing these instances to a higher level and shrinking the set of instances the benchmark can evaluate. (3)~We require all labeling sessions to be conducted offline without internet access. Since ReasonText is built on the publicly available RealSR-v3~\cite{realsrv3} and DRealSR~\cite{drealsr}, searching the web with an LR patch could retrieve its HR counterpart and expose the transcript. An instance resolved this way would be recorded as readable even when no cue in the image supports it, mixing looked-up text into the reasoning-required set.

\paragraph{Cue aggregation in Stage 2.}
A reasoning-required instance can be read through more than one route. For example, one annotator may identify a text from a repeated word elsewhere in the patch, while another identifies the same text from a brand they recognize. We therefore do not assign a single cue to an instance. Instead, we take the union of the cues selected by the annotators who identified the instance.

\paragraph{Level~3 text instances.}
Level~3 is defined by what our annotators could not read, not by what is inherently unreadable. A larger annotator pool or annotators with specific domain knowledge might recover text that no one in our pool could. We therefore release Level~3 instances with their polygons and transcripts, so that future work can revisit their levels with different annotator populations.

\paragraph{Statistics.}
ReasonText contains $513$ LR--HR pairs with $3{,}985$ annotated text instances, of which the $2{,}463$ Level~1 and Level~2 instances are used for main experiments. The instances are distributed as follows:

\begin{center}
\begin{tabular}{lrr}
\toprule
Level & \# instances & \% \\
\midrule
Level~1 (locally recoverable) & $1{,}754$ & $44.0$ \\
Level~2 (reasoning-required) & $709$ & $17.8$ \\
Level~3 (unrecoverable) & $1{,}522$ & $38.2$ \\
\midrule
Total & $3{,}985$ & $100.0$ \\
\bottomrule
\end{tabular}
\end{center}

\noindent
Among the $709$ Level~2 instances, $414$ are labeled with prior knowledge, $288$ with image context, and $246$ with logical patterns. These counts sum to more than $709$ because an instance carries the union of the cues selected by the annotators who identified it.

\begin{table}[t]
\centering
\setlength{\aboverulesep}{0pt}\setlength{\belowrulesep}{0pt}\renewcommand{\arraystretch}{1.2}
\caption{\textbf{Reliability of the ReasonText difficulty levels.} (a) Agreement, the fraction of annotators who attempted an instance and read it correctly, for the Level~1 and Level~2 instances assigned under our rule. The rows give the share of instances in each agreement range, followed by the mean and median. (b) Level counts re-derived by requiring at least one, at least half, more than half, or all of the attempting annotators to read an instance correctly. Agreement is high at both levels, and Level~2 does not shrink under stricter thresholds but grows from $709$ to $885$ under unanimity.
}
\label{tab:supp_tab_bench_stat}
\begin{subtable}[t]{0.46\textwidth}\centering
\caption{Inter-annotator agreement}\label{tab:agreement}
\begin{tabular}{l|cc}
\toprule
Agreement & L1 ($N{=}1754$) & L2 ($N{=}709$) \\
\midrule
all correct & 71.9\% & 73.6\% \\
50--99\% & 24.1\% & 19.5\% \\
$<$50\% & 3.9\% & 6.9\% \\
\midrule
mean & 89.3\% & 87.5\% \\
median & 100\% & 100\% \\
\bottomrule
\end{tabular}
\end{subtable}
\hfill
\begin{subtable}[t]{0.5\textwidth}\centering
\caption{Level counts by threshold}\label{tab:level_thresholds}
\begin{tabular}{l|ccc}
\toprule
Threshold & L1 & L2 & L3 \\
\midrule
$\geq$1 correct (ours) & 1754 & 709 & 1522 \\
majority ($\geq$50\%) & 1685 & 717 & 1583 \\
strict ($>$50\%) & 1571 & 737 & 1677 \\
unanimous ($100\%$) & 1262 & 885 & 1838 \\
\bottomrule
\end{tabular}
\end{subtable}
\end{table}

\paragraph{Robustness of the level assignment.}
We assign a level when any of the five annotators identifies an instance. Each instance has a GT transcript from its HR image, so what a level should capture is whether the text is recoverable at all, not how many annotators happen to recover it. Annotators differ in attention and in the prior knowledge they bring, and one correct reading is enough to establish that a route to the text exists. A majority rule would instead push text recovered by only some annotators to a harder level, making it appear harder to recover than it is.

Table~\ref{tab:supp_tab_bench_stat}(a) reports agreement, defined for each instance as the fraction of attempting annotators whose transcript matches the GT transcript, for the levels assigned under our rule. We exclude those who skipped, since they are instructed to skip whenever they are unsure. Agreement averages $89.3\%$ for Level~1 and $87.5\%$ for Level~2, with a median of $100\%$ for both, and all attempting annotators read the text correctly for $71.9\%$ of Level~1 and $73.6\%$ of Level~2 instances. Agreement is thus high at both levels, and the assigned levels are stable.

Table~\ref{tab:supp_tab_bench_stat}(b) re-derives the levels by re-running the two-stage assignment under stricter thresholds, which count an instance as identified only when at least half, more than half, or all of its attempting annotators read it correctly. Under the majority rule, the share of each level changes by less than two percentage points. Even under unanimity, Level~2 does not shrink but grows from $709$ to $885$, as instances that leave Level~1 are re-examined in Stage~2 and recovered there with the surrounding context. The reasoning-required set therefore persists under every threshold we tested.

Re-evaluating the SR models on the levels re-derived under the majority rule leads to the same conclusions (Table~\ref{tab:supp_tab_bench_table}). SR models still fail on reasoning-required text: GTTCA drops sharply from Level~1 to Level~2 for every model, and the lowest Level~1 score ($46.4\%$) still far exceeds the highest Level~2 score ($26.9\%$). RTC is effective as well, improving all six text-conditioned SR models at both levels, with larger gains on Level 2 for most models. Moreover, the real-world SR model DiT4SR with RTC again outperforms the TAISR model UniT with RTC.

\begin{table}[t]
\centering
\setlength{\aboverulesep}{0pt}\setlength{\belowrulesep}{0pt}\renewcommand{\arraystretch}{1.2}
\caption{\textbf{The conclusions hold under the majority rule.} GTTCA on ReasonText with the levels re-derived under the majority ($\geq$50\%) rule of Table~\ref{tab:supp_tab_bench_stat} (b), which gives $1{,}685$ Level~1 and $717$ Level~2 instances. All twelve models drop sharply on Level~2, RTC improves all six text-conditioned SR models, and DiT4SR\,+\,RTC outperforms UniT\,+\,RTC. \textcolor{ImpUp}{Blue} marks an improvement over the baseline.}
\label{tab:supp_tab_bench_table}
\begin{tabular}{cl|ccc|ccc}
\toprule
\multirow{2}{*}{Group} & \multirow{2}{*}{Model} & \multicolumn{3}{c|}{Baseline} & \multicolumn{3}{c}{+ RTC} \\
\cmidrule(lr){3-5} \cmidrule(lr){6-8}
& & Overall$\uparrow$ & Level~1$\uparrow$ & Level~2$\uparrow$ & Overall$\uparrow$ & Level~1$\uparrow$ & Level~2$\uparrow$ \\
\midrule
\multirow{2}{*}{A} & Real-ESRGAN & 41.0 & 53.9 & 10.7 & -- & -- & -- \\
 & SwinIR & 46.4 & 60.1 & 14.2 & -- & -- & -- \\
\midrule
\multirow{4}{*}{B} & DiffBIR & 55.7 & 70.9 & 20.2 & -- & -- & -- \\
 & SUPIR & 55.8 & 69.3 & 24.1 & \textcolor{ImpUp}{67.8} & \textcolor{ImpUp}{79.5} & \textcolor{ImpUp}{40.2} \\
 & FaithDiff & 52.5 & 65.9 & 20.8 & \textcolor{ImpUp}{61.6} & \textcolor{ImpUp}{74.8} & \textcolor{ImpUp}{30.7} \\
 & DiT4SR & 60.0 & 74.1 & 26.9 & \textcolor{ImpUp}{74.0} & \textcolor{ImpUp}{85.0} & \textcolor{ImpUp}{48.1} \\
\midrule
\multirow{3}{*}{C} & ResShift & 48.4 & 61.8 & 17.0 & -- & -- & -- \\
 & SinSR & 43.8 & 56.9 & 13.1 & -- & -- & -- \\
 & OSEDiff & 45.2 & 59.1 & 12.7 & \textcolor{ImpUp}{49.9} & \textcolor{ImpUp}{65.0} & \textcolor{ImpUp}{14.5} \\
\midrule
D & PURE & 35.0 & 46.4 & 8.2 & -- & -- & -- \\
\midrule
\multirow{2}{*}{E} & TeReDiff & 52.0 & 65.9 & 19.2 & \textcolor{ImpUp}{57.2} & \textcolor{ImpUp}{71.0} & \textcolor{ImpUp}{24.8} \\
 & UniT & 61.8 & 76.9 & 26.2 & \textcolor{ImpUp}{71.2} & \textcolor{ImpUp}{83.1} & \textcolor{ImpUp}{43.2} \\
\bottomrule
\end{tabular}
\end{table}

\subsection{GTTCA Metric}
\label{sec:supp_metric}

\begin{figure*}[htbp]
    \centering
    \includegraphics[width=\linewidth]{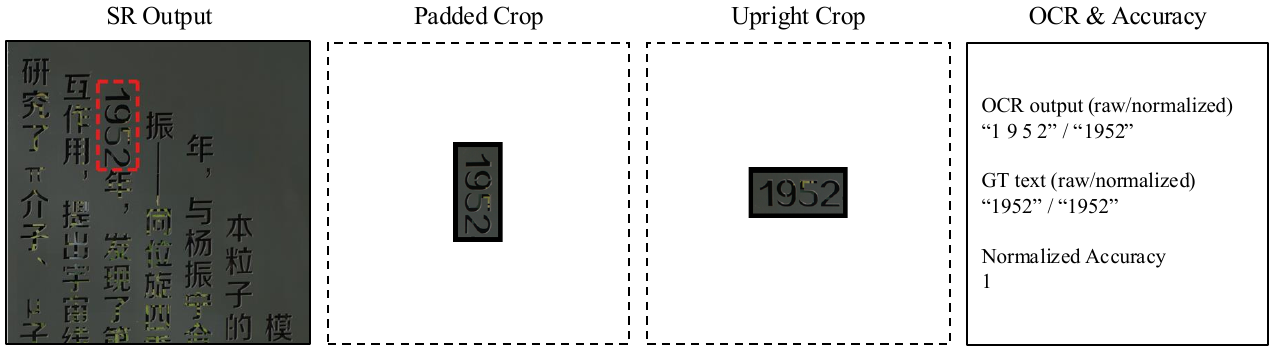}

    \caption{\textbf{Computing GTTCA on a restored text instance.} Each text instance is cropped around its GT polygon with padding, rotated and flipped with its GT orientation, and recognized by an OCR model. The recognized text is compared with the GT transcript under normalized exact match.}

    \label{fig:supp_fig_metric}
\end{figure*}

Figure~\ref{fig:supp_fig_metric} illustrates the four steps of GTTCA for a single text instance. We further validate the design of GTTCA (Section~\ref{subsec:metric}) with two experiments.

\begin{table}[t]
\centering
\setlength{\aboverulesep}{0pt}\setlength{\belowrulesep}{0pt}\renewcommand{\arraystretch}{1.2}
\caption{\textbf{Both GT polygons and GT orientations are needed for GTTCA to reflect restoration alone.} All three protocols share the same OCR backbone~\cite{paddleocrvl15}. GTTCA uses both GT polygons and GT orientations, w/o RF uses GT polygons without rotation and flipping, and OCR detects and recognizes text end-to-end without either. The HR oracle row shows the ceiling of each protocol on images that need no restoration. Excluding the HR oracle, \textbf{bold} marks the best score in each column, and \underline{underline} marks the second-best.}
\label{tab:supp_tab_metric_ablation}
\small
\setlength{\tabcolsep}{4pt}
\begin{tabular}{ll|ccc|ccc|ccc}
\toprule
\multirow{2}{*}{Group} & \multirow{2}{*}{Model} & \multicolumn{3}{c|}{Overall} & \multicolumn{3}{c|}{Level 1} & \multicolumn{3}{c}{Level 2} \\
\cmidrule(lr){3-5} \cmidrule(lr){6-8} \cmidrule(lr){9-11}
& & GTTCA & w/o RF & OCR & GTTCA & w/o RF & OCR & GTTCA & w/o RF & OCR \\
\midrule
\rowcolor{cyan!20}
\multicolumn{2}{c|}{HR (oracle)} & 94.7 & 92.7 & 29.0 & 96.3 & 94.7 & 30.0 & 90.7 & 87.6 & 26.7 \\
\midrule
\multirow{2}{*}{A} & Real-ESRGAN & 40.3 & 38.2 & 12.0 & 53.0 & 50.7 & 16.6 & 9.0 & 7.3 & 0.6 \\
 & SwinIR & 45.7 & 43.3 & 13.9 & 59.4 & 57.0 & 18.6 & 11.7 & 9.3 & 2.3 \\
\midrule
\multirow{4}{*}{B} & DiffBIR & 55.0 & 53.5 & 15.1 & 69.6 & 67.8 & 20.0 & 18.8 & 17.9 & 3.0 \\
 & SUPIR & 55.3 & 54.0 & 16.8 & 68.5 & 67.1 & \underline{21.6} & 22.6 & 21.4 & \underline{4.9} \\
 & FaithDiff & 52.0 & 50.4 & 15.6 & 65.2 & 63.4 & 20.1 & 19.5 & 18.2 & 4.4 \\
 & DiT4SR & \underline{59.2} & \underline{57.8} & \underline{17.0} & \underline{73.1} & \underline{71.7} & 21.3 & \textbf{24.5} & \underline{23.4} & \textbf{6.2} \\
\midrule
\multirow{3}{*}{C} & ResShift & 47.7 & 45.5 & 14.3 & 61.0 & 58.7 & 19.6 & 14.8 & 13.0 & 1.4 \\
 & SinSR & 43.1 & 41.1 & 12.8 & 56.0 & 53.6 & 17.5 & 11.3 & 10.2 & 1.1 \\
 & OSEDiff & 44.6 & 43.4 & 12.5 & 58.4 & 57.2 & 17.1 & 10.4 & 9.0 & 1.3 \\
\midrule
D & PURE & 34.4 & 33.2 & 10.4 & 45.7 & 44.1 & 14.2 & 6.3 & 6.1 & 1.1 \\
\midrule
\multirow{2}{*}{E} & TeReDiff & 51.2 & 49.8 & 16.1 & 65.1 & 63.6 & 21.6 & 16.9 & 15.8 & 2.5 \\
 & UniT & \textbf{61.4} & \textbf{59.7} & \textbf{17.2} & \textbf{76.5} & \textbf{74.3} & \textbf{22.4} & \underline{24.1} & \textbf{23.7} & 4.2 \\
\bottomrule
\end{tabular}
\end{table}

\paragraph{Ablation of the GTTCA protocol.}
GTTCA differs from an end-to-end text spotting protocol in two ways: it uses GT polygons instead of a text detector, and it rotates and flips each crop with the GT orientation so that the text reads upright. Table~\ref{tab:supp_tab_metric_ablation} removes these two components one after another, with all protocols using PaddleOCR-VL-1.5~\cite{paddleocrvl15} so that any change comes from the protocol rather than the recognizer. The \textbf{w/o RF} (without rotation and flipping) protocol keeps the GT polygons but recognizes each crop in its original orientation. The \textbf{OCR} protocol additionally drops the GT polygons and uses PaddleOCR-VL-1.5 in its end-to-end mode, which detects the text and then recognizes it.

Both components remove a factor that has nothing to do with restoration quality. Without GT polygons, the score depends on whether the detector finds the text: the \textbf{OCR} protocol reaches only $29.0\%$ on the HR images, which need no restoration at all, and compresses the twelve models into a seven-point range ($10.4\%$ to $17.2\%$). The effect is most severe on Level~2, where eight of the twelve models score $3.0\%$ or lower, leaving no room to compare them on reasoning-required text. Without rotation and flipping, the score depends on how the text happens to be oriented in the image. The \textbf{w/o RF} protocol lowers overall GTTCA by $1.2$ to $2.4$ points for every model and by $2.0$ points for the HR oracle ($94.7\%$ to $92.7\%$), and the loss is uneven enough to change the ranking, even swapping DiT4SR and UniT at the top of Level 2.

\begin{table}[t]
\centering
\setlength{\aboverulesep}{0pt}\setlength{\belowrulesep}{0pt}\renewcommand{\arraystretch}{1.2}
\caption{\textbf{GTTCA under different OCR recognizers.} Overall GTTCA recomputed with PaddleOCR-VL-1.5 (\textbf{Paddle}), ASTER, and MORAN. The weaker recognizers lower every score by a similar proportion, so the ranking of the models barely changes (Table~\ref{tab:supp_tab_metric_rank}). In the + RTC columns, \textcolor{ImpUp}{blue} marks an improvement over the baseline under the same recognizer.}
\label{tab:supp_tab_metric_robust}
\begin{tabular}{cl|ccc|ccc}
\toprule
\multirow{2}{*}{Group} & \multirow{2}{*}{Model} & \multicolumn{3}{c|}{Baseline (Overall)} & \multicolumn{3}{c}{+ RTC (Overall)} \\
\cmidrule(lr){3-5} \cmidrule(lr){6-8}
& & Paddle & ASTER & MORAN & Paddle & ASTER & MORAN \\
\midrule
\multirow{2}{*}{A} & Real-ESRGAN & 40.3 & 25.8 & 26.8 & -- & -- & -- \\
 & SwinIR & 45.7 & 30.9 & 30.8 & -- & -- & -- \\
\midrule
\multirow{4}{*}{B} & DiffBIR & 55.0 & 38.0 & 40.1 & -- & -- & -- \\
 & SUPIR & 55.3 & 38.3 & 41.7 & \textcolor{ImpUp}{67.3} & \textcolor{ImpUp}{49.2} & \textcolor{ImpUp}{50.9} \\
 & FaithDiff & 52.0 & 36.4 & 38.3 & \textcolor{ImpUp}{61.0} & \textcolor{ImpUp}{43.6} & \textcolor{ImpUp}{45.0} \\
 & DiT4SR & 59.2 & 41.7 & 44.8 & \textcolor{ImpUp}{73.6} & \textcolor{ImpUp}{54.7} & \textcolor{ImpUp}{56.1} \\
\midrule
\multirow{3}{*}{C} & ResShift & 47.7 & 33.8 & 33.2 & -- & -- & -- \\
 & SinSR & 43.1 & 31.9 & 31.5 & -- & -- & -- \\
 & OSEDiff & 44.6 & 29.7 & 32.7 & \textcolor{ImpUp}{49.4} & \textcolor{ImpUp}{33.0} & \textcolor{ImpUp}{35.8} \\
\midrule
D & PURE & 34.4 & 24.8 & 25.9 & -- & -- & -- \\
\midrule
\multirow{2}{*}{E} & TeReDiff & 51.2 & 35.0 & 37.7 & \textcolor{ImpUp}{56.6} & \textcolor{ImpUp}{38.8} & \textcolor{ImpUp}{41.0} \\
 & UniT & 61.4 & 43.7 & 46.3 & \textcolor{ImpUp}{71.0} & \textcolor{ImpUp}{49.7} & \textcolor{ImpUp}{52.3} \\
\bottomrule
\end{tabular}%
\end{table}

\begin{table}[t]
\centering
\setlength{\aboverulesep}{0pt}\setlength{\belowrulesep}{0pt}\renewcommand{\arraystretch}{1.2}
\caption{\textbf{Ranking agreement across recognizers.} Kendall $\tau$ and Spearman $\rho$ between the model ranking under PaddleOCR-VL-1.5 and those under ASTER and MORAN, computed on Overall GTTCA in Table~\ref{tab:supp_tab_metric_robust}. $1.0$ denotes an identical ranking. Both measures stay near $1.0$, so GTTCA ranks the SR models in nearly the same order regardless of which recognizer reads the output.}
\label{tab:supp_tab_metric_rank}
\begin{tabular}{l|cc|cc}
\toprule
Ranking (Overall GTTCA) & $\tau$ vs ASTER & $\tau$ vs MORAN & $\rho$ vs ASTER & $\rho$ vs MORAN \\
\midrule
12 SR models & 0.94 & 0.94 & 0.98 & 0.98 \\
+ RTC (6 SR models) & 1.00 & 1.00 & 1.00 & 1.00 \\
\bottomrule
\end{tabular}
\end{table}

\paragraph{Robustness to the OCR model.}
GTTCA reads restored text with an OCR model, which raises the question of whether the results would hold with a different recognizer. We therefore recompute GTTCA with ASTER~\cite{aster} and MORAN~\cite{moran}, two recognizers standard in STISR evaluation, and report the results in Table~\ref{tab:supp_tab_metric_robust}.

Since both recognizers are older and weaker than PaddleOCR-VL-1.5~\cite{paddleocrvl15}, absolute scores fall as expected. However, the drop is similar in proportion ($24\%$ to $36\%$), so the ranking is nearly preserved, with Kendall $\tau \geq 0.94$ and Spearman $\rho \geq 0.98$ against PaddleOCR-VL-1.5 for both recognizers (Table~\ref{tab:supp_tab_metric_rank}). GTTCA is therefore robust to the choice of recognizer, since a weaker recognizer lowers every score but barely changes which models restore text better.

\subsection{SR Models Fail on Reasoning-Required Text}
\label{sec:supp_failure}

\begin{figure*}[htbp]
    \centering
    \includegraphics[width=\linewidth]{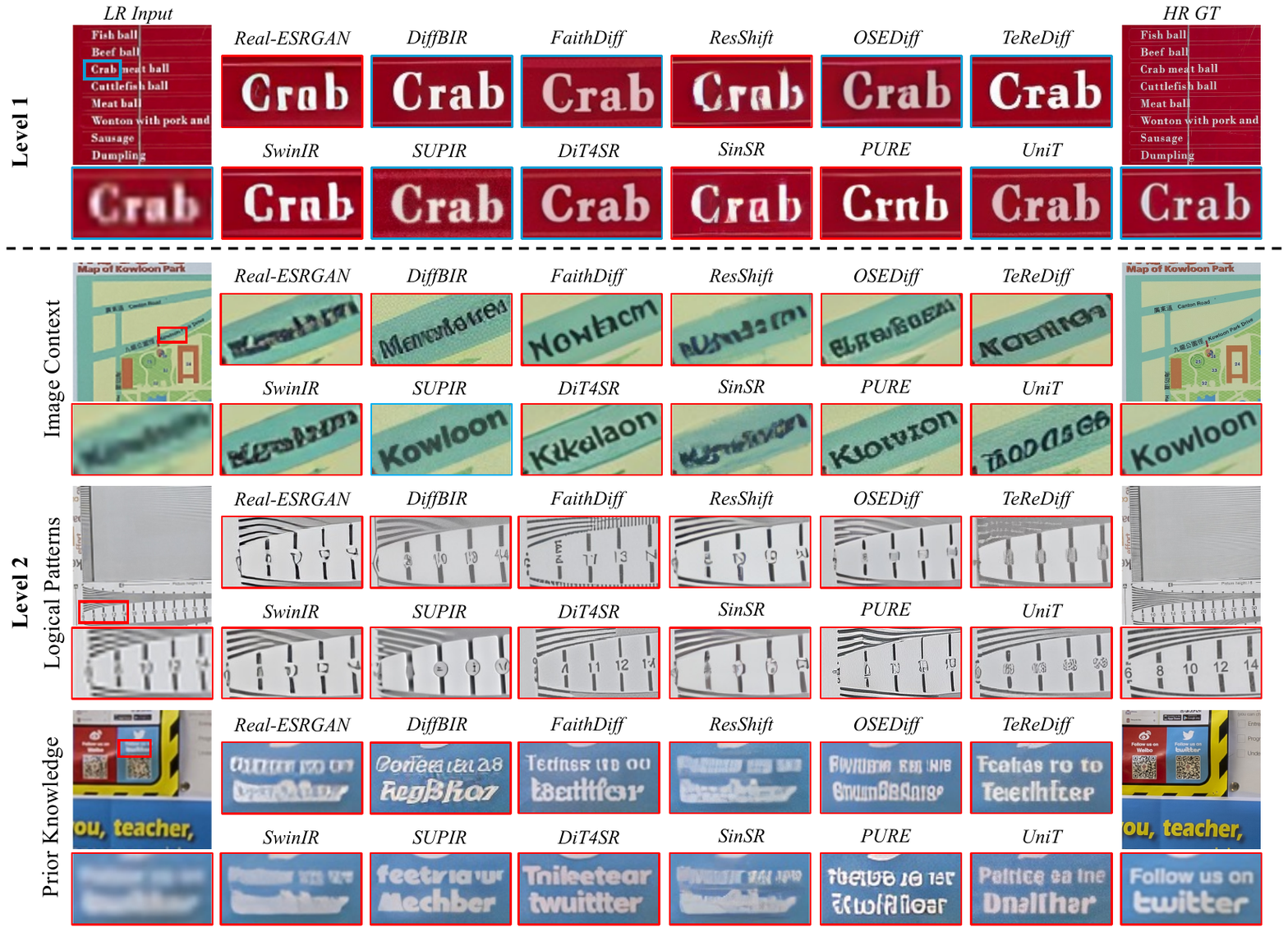}
    \caption{\textbf{SR models fail on reasoning-required text.} Outputs of the twelve SR models on one Level~1 (locally recoverable) instance and three Level~2 (reasoning-required) instances, one for each cue. \textcolor{ImpUp}{Blue} and \textcolor{ImpDown}{red} borders on the SR outputs mark correct and incorrect restorations. On Level~1, most models restore ``Crab'' correctly. On Level~2, most models fail to restore the text.} 
    \label{supp_fig:supp_fig_failure}
\end{figure*}

Figure~\ref{supp_fig:supp_fig_failure} extends Section~\ref{subsec:failure} with qualitative comparisons of the twelve SR models on one Level~1 instance and three Level~2 instances, one for each cue. On Level~1, most models restore ``Crab'' correctly. On Level~2, however, most models fail regardless of architecture or text-awareness, visually echoing the GTTCA drop reported in Table~\ref{tab:tab_failure}.

\section{Additional Details on Reasoning Transfer via Captioning}
\label{sec:supp_rtc}

\subsection{Observation 1: Ground-Truth Transcripts Improve Text Restoration}
\label{sec:supp_rtc_injection}

\begin{table}[t]
\centering
\setlength{\aboverulesep}{0pt}
\setlength{\belowrulesep}{0pt}
\renewcommand{\arraystretch}{1.25}
\caption{\textbf{The natural-language caption restores text best among the three caption formats.} Each text-conditioned SR model is evaluated with its original caption and with the GT transcripts of its Level~1 and Level~2 instances injected in three formats. Within each model block, \textbf{bold} marks the best value for each metric, and \textcolor{ImpUp}{blue}/\textcolor{ImpDown}{red} indicate improvement/decline over the original caption. The HR oracle row reports each metric on the HR images for reference. $\uparrow$($\downarrow$): higher (lower) is better.}
\label{tab:supp_tab_rtc_capform}
\resizebox{\textwidth}{!}{%
\begin{tabular}{cl|cc|cc|ccc|ccc}
\toprule
\multirow{3}{*}{Group} & \multirow{3}{*}{Model} & \multicolumn{2}{c|}{\multirow{2}{*}{Fidelity}} & \multicolumn{5}{c|}{Perceptual} & \multicolumn{3}{c}{Text Restoration} \\
\cmidrule(lr){5-9} \cmidrule(lr){10-12}
& & \multicolumn{2}{c|}{} & \multicolumn{2}{c|}{Reference} & \multicolumn{3}{c|}{No Reference} & \multicolumn{3}{c}{GTTCA} \\
\cmidrule(lr){3-4} \cmidrule(lr){5-6} \cmidrule(lr){7-9} \cmidrule(lr){10-12}
& & PSNR-Y$\uparrow$ & SSIM-Y$\uparrow$ & LPIPS$\downarrow$ & DISTS$\downarrow$ & MANIQA$\uparrow$ & MUSIQ$\uparrow$ & CLIP-IQA$\uparrow$ & Overall$\uparrow$ & Level~1$\uparrow$ & Level~2$\uparrow$ \\
\midrule
\rowcolor{cyan!20}
\multicolumn{2}{c|}{HR (oracle)} & 80.0000 & 1.0000 & 0.0000 & 0.0000 & 0.4261 & 66.2367 & 0.5403 & 94.6813 & 96.2942 & 90.6911 \\
\midrule
\multirow{12}{*}{B} & SUPIR & 24.2971 & \textbf{0.7098} & \textbf{0.2622} & \textbf{0.1947} & \textbf{0.4608} & \textbf{63.5998} & \textbf{0.5922} & 55.2984 & 68.5291 & 22.5670 \\
 & \quad + tag style GT (space) & \textcolor{ImpUp}{24.7121} & \textcolor{ImpDown}{0.7014} & \textcolor{ImpDown}{0.2840} & \textcolor{ImpDown}{0.2013} & \textcolor{ImpDown}{0.4332} & \textcolor{ImpDown}{60.6653} & \textcolor{ImpDown}{0.5659} & \textcolor{ImpUp}{64.9695} & \textcolor{ImpUp}{77.9590} & \textcolor{ImpUp}{32.8350} \\
 & \quad + tag style GT (comma) & \textcolor{ImpUp}{\textbf{24.7246}} & \textcolor{ImpDown}{0.7006} & \textcolor{ImpDown}{0.2852} & \textcolor{ImpDown}{0.2024} & \textcolor{ImpDown}{0.4291} & \textcolor{ImpDown}{60.2502} & \textcolor{ImpDown}{0.5620} & \textcolor{ImpUp}{64.6935} & \textcolor{ImpUp}{78.1984} & \textcolor{ImpUp}{31.2835} \\
 & \quad + caption style GT & \textcolor{ImpUp}{24.5912} & \textcolor{ImpDown}{0.7038} & \textcolor{ImpDown}{0.2732} & \textcolor{ImpDown}{0.1964} & \textcolor{ImpDown}{0.4349} & \textcolor{ImpDown}{60.9064} & \textcolor{ImpDown}{0.5714} & \textcolor{ImpUp}{\textbf{69.7686}} & \textcolor{ImpUp}{\textbf{81.6534}} & \textcolor{ImpUp}{\textbf{40.3667}} \\
\cmidrule(l){2-12}
 & FaithDiff & 24.2038 & 0.7363 & 0.2061 & 0.1623 & \textbf{0.5095} & \textbf{71.6540} & \textbf{0.5821} & 52.0097 & 65.1653 & 19.4640 \\
 & \quad + tag style GT (space) & \textcolor{ImpUp}{24.3528} & \textcolor{ImpUp}{0.7407} & \textcolor{ImpUp}{0.2018} & \textcolor{ImpUp}{0.1595} & \textcolor{ImpDown}{0.4871} & \textcolor{ImpDown}{71.2038} & \textcolor{ImpDown}{0.5758} & \textcolor{ImpUp}{61.5510} & \textcolor{ImpUp}{74.6522} & \textcolor{ImpUp}{29.1396} \\
 & \quad + tag style GT (comma) & \textcolor{ImpUp}{24.3674} & \textcolor{ImpUp}{0.7415} & \textcolor{ImpUp}{0.2021} & \textcolor{ImpUp}{0.1600} & \textcolor{ImpDown}{0.4926} & \textcolor{ImpDown}{71.3235} & \textcolor{ImpDown}{0.5777} & \textcolor{ImpUp}{61.2667} & \textcolor{ImpUp}{74.7320} & \textcolor{ImpUp}{27.9549} \\
 & \quad + caption style GT & \textcolor{ImpUp}{\textbf{24.4677}} & \textcolor{ImpUp}{\textbf{0.7467}} & \textcolor{ImpUp}{\textbf{0.1962}} & \textcolor{ImpUp}{\textbf{0.1567}} & \textcolor{ImpDown}{0.4951} & \textcolor{ImpDown}{71.4650} & \textcolor{ImpDown}{0.5739} & \textcolor{ImpUp}{\textbf{62.4523}} & \textcolor{ImpUp}{\textbf{75.3250}} & \textcolor{ImpUp}{\textbf{30.6065}} \\
\cmidrule(l){2-12}
 & DiT4SR & 22.6539 & 0.6933 & 0.2477 & 0.1830 & 0.4973 & 70.8502 & \textbf{0.6074} & 59.1555 & 73.1471 & 24.5416 \\
 & \quad + tag style GT (space) & \textcolor{ImpUp}{22.8881} & \textcolor{ImpUp}{0.7122} & \textcolor{ImpUp}{0.2360} & \textcolor{ImpUp}{0.1814} & \textcolor{ImpUp}{\textbf{0.5015}} & \textcolor{ImpUp}{\textbf{71.1707}} & \textcolor{ImpDown}{0.5979} & \textcolor{ImpUp}{69.9959} & \textcolor{ImpUp}{82.5314} & \textcolor{ImpUp}{38.9845} \\
 & \quad + tag style GT (comma) & \textcolor{ImpUp}{22.9431} & \textcolor{ImpUp}{0.7143} & \textcolor{ImpUp}{0.2339} & \textcolor{ImpUp}{0.1805} & \textcolor{ImpUp}{0.4983} & \textcolor{ImpUp}{71.0330} & \textcolor{ImpDown}{0.5951} & \textcolor{ImpUp}{70.1340} & \textcolor{ImpUp}{82.8050} & \textcolor{ImpUp}{38.7870} \\
 & \quad + caption style GT & \textcolor{ImpUp}{\textbf{23.0546}} & \textcolor{ImpUp}{\textbf{0.7182}} & \textcolor{ImpUp}{\textbf{0.2317}} & \textcolor{ImpUp}{\textbf{0.1797}} & \textcolor{ImpDown}{0.4944} & \textcolor{ImpDown}{70.8355} & \textcolor{ImpDown}{0.5932} & \textcolor{ImpUp}{\textbf{74.1048}} & \textcolor{ImpUp}{\textbf{85.1767}} & \textcolor{ImpUp}{\textbf{46.7137}} \\
\midrule
\multirow{4}{*}{C} & OSEDiff & \textbf{23.9596} & 0.7533 & \textbf{0.2147} & \textbf{0.1666} & 0.4803 & 71.1559 & 0.5672 & 44.6204 & 58.4379 & 10.4372 \\
 & \quad + tag style GT (space) & \textcolor{ImpDown}{23.9309} & \textcolor{ImpDown}{0.7531} & \textcolor{ImpDown}{0.2170} & \textcolor{ImpDown}{0.1679} & \textcolor{ImpDown}{0.4791} & \textcolor{ImpUp}{\textbf{71.1938}} & \textcolor{ImpUp}{\textbf{0.5718}} & \textcolor{ImpUp}{49.4762} & \textcolor{ImpUp}{64.8461} & \textcolor{ImpUp}{11.4528} \\
 & \quad + tag style GT (comma) & \textcolor{ImpDown}{23.9462} & \textcolor{ImpDown}{0.7532} & \textcolor{ImpDown}{0.2166} & \textcolor{ImpDown}{0.1675} & \textcolor{ImpDown}{0.4760} & \textcolor{ImpDown}{71.0987} & \textcolor{ImpUp}{0.5716} & \textcolor{ImpUp}{49.7199} & \textcolor{ImpUp}{64.6750} & \textcolor{ImpUp}{12.7221} \\
 & \quad + caption style GT & \textcolor{ImpDown}{23.9394} & \textcolor{ImpUp}{\textbf{0.7546}} & \textcolor{ImpDown}{0.2179} & \textcolor{ImpDown}{0.1695} & \textcolor{ImpUp}{\textbf{0.4811}} & \textcolor{ImpDown}{71.1110} & \textcolor{ImpUp}{0.5676} & \textcolor{ImpUp}{\textbf{51.4657}} & \textcolor{ImpUp}{\textbf{66.4082}} & \textcolor{ImpUp}{\textbf{14.4993}} \\
\midrule
\multirow{8}{*}{E} & TeReDiff & \textbf{23.5188} & \textbf{0.7233} & \textbf{0.2362} & \textbf{0.1796} & 0.5638 & 70.5181 & 0.6183 & 51.2383 & 65.1083 & 16.9252 \\
 & \quad + tag style GT (space) & \textcolor{ImpDown}{23.2760} & \textcolor{ImpDown}{0.7188} & \textcolor{ImpDown}{0.2393} & \textcolor{ImpDown}{0.1820} & \textcolor{ImpDown}{0.5608} & \textcolor{ImpUp}{70.7581} & \textcolor{ImpDown}{0.6166} & \textcolor{ImpUp}{55.3553} & \textcolor{ImpUp}{69.1448} & \textcolor{ImpUp}{21.2412} \\
 & \quad + tag style GT (comma) & \textcolor{ImpDown}{23.3000} & \textcolor{ImpDown}{0.7200} & \textcolor{ImpDown}{0.2382} & \textcolor{ImpDown}{0.1814} & \textcolor{ImpDown}{0.5611} & \textcolor{ImpUp}{70.8026} & \textcolor{ImpDown}{0.6142} & \textcolor{ImpUp}{56.6301} & \textcolor{ImpUp}{70.0912} & \textcolor{ImpUp}{23.3286} \\
 & \quad + caption style GT & \textcolor{ImpDown}{23.1785} & \textcolor{ImpDown}{0.7176} & \textcolor{ImpDown}{0.2388} & \textcolor{ImpDown}{0.1812} & \textcolor{ImpUp}{\textbf{0.5694}} & \textcolor{ImpUp}{\textbf{71.0449}} & \textcolor{ImpUp}{\textbf{0.6251}} & \textcolor{ImpUp}{\textbf{59.5128}} & \textcolor{ImpUp}{\textbf{72.7708}} & \textcolor{ImpUp}{\textbf{26.7137}} \\
\cmidrule(l){2-12}
 & UniT & 22.7672 & 0.6503 & 0.2814 & 0.2072 & \textbf{0.5159} & \textbf{68.7382} & \textbf{0.6315} & 61.3886 & 76.4538 & 24.1185 \\
 & \quad + tag style GT (space) & \textcolor{ImpUp}{23.0685} & \textcolor{ImpUp}{0.6699} & \textcolor{ImpUp}{0.2722} & \textcolor{ImpUp}{0.2004} & \textcolor{ImpDown}{0.5042} & \textcolor{ImpDown}{68.3784} & \textcolor{ImpDown}{0.6182} & \textcolor{ImpUp}{64.5635} & \textcolor{ImpUp}{78.7001} & \textcolor{ImpUp}{29.5910} \\
 & \quad + tag style GT (comma) & \textcolor{ImpUp}{23.0727} & \textcolor{ImpUp}{0.6701} & \textcolor{ImpUp}{0.2721} & \textcolor{ImpUp}{0.2003} & \textcolor{ImpDown}{0.5039} & \textcolor{ImpDown}{68.3584} & \textcolor{ImpDown}{0.6181} & \textcolor{ImpUp}{64.4417} & \textcolor{ImpUp}{78.5063} & \textcolor{ImpUp}{29.6474} \\
 & \quad + caption style GT & \textcolor{ImpUp}{\textbf{23.1021}} & \textcolor{ImpUp}{\textbf{0.6725}} & \textcolor{ImpUp}{\textbf{0.2694}} & \textcolor{ImpUp}{\textbf{0.1994}} & \textcolor{ImpDown}{0.5024} & \textcolor{ImpDown}{68.3612} & \textcolor{ImpDown}{0.6169} & \textcolor{ImpUp}{\textbf{70.6618}} & \textcolor{ImpUp}{\textbf{82.8848}} & \textcolor{ImpUp}{\textbf{40.4231}} \\
\bottomrule
\end{tabular}%
}
\end{table}

This section provides two ablations on the GT transcript caption of Section~\ref{sec:injection}, one on its format and one on adding a scene description.

\paragraph{Caption format ablation.}
Text-conditioned SR models differ in the caption format they are trained with. It is therefore unclear whether the GT transcripts should be injected in each model's training-time format or in a single unified format. We compare three formats that mirror the conventions of existing SR models, namely (1) tag style GT (space), which separates the transcripts with spaces~\cite{unit} (\texttt{"$t_1$ $t_2$ $\cdots$ $t_n$"}), (2) tag style GT (comma), which separates them with commas~\cite{osediff} (\texttt{"$t_1$, $t_2$, $\cdots$, $t_n$"}), and (3) caption style GT, which lists them in a natural-language caption (\texttt{"The image with the text \textquotedblleft $t_1$\textquotedblright, \textquotedblleft $t_2$\textquotedblright, $\cdots$, \textquotedblleft $t_n$\textquotedblright."}).

Table~\ref{tab:supp_tab_rtc_capform} shows that GT transcripts improve GTTCA over the original caption for every model, regardless of the format. Among the three formats, the natural-language caption achieves the best GTTCA for every model on both levels, even for UniT and OSEDiff, whose own conventions are tag style. We hypothesize that this is because the text-to-image models underlying these SR models were pretrained on natural-language captions, even though the SR models were later fine-tuned with various caption formats. These results support our choice in Section~\ref{sec:injection} to inject the GT transcripts as a natural-language caption.

\begin{table}[t]
\centering
\setlength{\aboverulesep}{0pt}
\setlength{\belowrulesep}{0pt}
\renewcommand{\arraystretch}{1.25}
\caption{\textbf{Adding a scene description to the caption-style GT largely recovers no-reference perceptual quality.} Each text-conditioned SR model is evaluated with its original caption, with the caption-style GT of its Level~1 and Level~2 instances, and with a Gemma 4 scene description prepended to that caption. Within each model block, \textbf{bold} marks the best value for each metric, and \textcolor{ImpUp}{blue}/\textcolor{ImpDown}{red} indicate improvement/decline over the original caption. The HR oracle row reports each metric on the HR images for reference. $\uparrow$($\downarrow$): higher (lower) is better.
}
\label{tab:supp_tab_rtc_ablation}
\resizebox{\textwidth}{!}{%
\begin{tabular}{cl|cc|cc|ccc|ccc}
\toprule
\multirow{3}{*}{Group} & \multirow{3}{*}{Model} & \multicolumn{2}{c|}{\multirow{2}{*}{Fidelity}} & \multicolumn{5}{c|}{Perceptual} & \multicolumn{3}{c}{Text Restoration} \\
\cmidrule(lr){5-9} \cmidrule(lr){10-12}
& & \multicolumn{2}{c|}{} & \multicolumn{2}{c|}{Reference} & \multicolumn{3}{c|}{No Reference} & \multicolumn{3}{c}{GTTCA} \\
\cmidrule(lr){3-4} \cmidrule(lr){5-6} \cmidrule(lr){7-9} \cmidrule(lr){10-12}
& & PSNR-Y$\uparrow$ & SSIM-Y$\uparrow$ & LPIPS$\downarrow$ & DISTS$\downarrow$ & MANIQA$\uparrow$ & MUSIQ$\uparrow$ & CLIP-IQA$\uparrow$ & Overall$\uparrow$ & Level~1$\uparrow$ & Level~2$\uparrow$ \\
\midrule
\rowcolor{cyan!20}
\multicolumn{2}{c|}{HR (oracle)} & 80.0000 & 1.0000 & 0.0000 & 0.0000 & 0.4261 & 66.2367 & 0.5403 & 94.6813 & 96.2942 & 90.6911 \\
\midrule
\multirow{9}{*}{B} & SUPIR & 24.2971 & \textbf{0.7098} & 0.2622 & 0.1947 & 0.4608 & 63.5998 & 0.5922 & 55.2984 & 68.5291 & 22.5670 \\
 & \quad + caption style GT & \textcolor{ImpUp}{\textbf{24.5912}} & \textcolor{ImpDown}{0.7038} & \textcolor{ImpDown}{0.2732} & \textcolor{ImpDown}{0.1964} & \textcolor{ImpDown}{0.4349} & \textcolor{ImpDown}{60.9064} & \textcolor{ImpDown}{0.5714} & \textcolor{ImpUp}{\textbf{69.7686}} & \textcolor{ImpUp}{\textbf{81.6534}} & \textcolor{ImpUp}{\textbf{40.3667}} \\
 & \quad + scene + caption style GT & \textcolor{ImpDown}{23.9171} & \textcolor{ImpDown}{0.7059} & \textcolor{ImpUp}{\textbf{0.2497}} & \textcolor{ImpUp}{\textbf{0.1828}} & \textcolor{ImpUp}{\textbf{0.5293}} & \textcolor{ImpUp}{\textbf{67.4045}} & \textcolor{ImpUp}{\textbf{0.6405}} & \textcolor{ImpUp}{67.9009} & \textcolor{ImpUp}{79.0878} & \textcolor{ImpUp}{40.2257} \\
\cmidrule(l){2-12}
 & FaithDiff & 24.2038 & 0.7363 & 0.2061 & 0.1623 & \textbf{0.5095} & \textbf{71.6540} & \textbf{0.5821} & 52.0097 & 65.1653 & 19.4640 \\
 & \quad + caption style GT & \textcolor{ImpUp}{\textbf{24.4677}} & \textcolor{ImpUp}{\textbf{0.7467}} & \textcolor{ImpUp}{\textbf{0.1962}} & \textcolor{ImpUp}{\textbf{0.1567}} & \textcolor{ImpDown}{0.4951} & \textcolor{ImpDown}{71.4650} & \textcolor{ImpDown}{0.5739} & \textcolor{ImpUp}{\textbf{62.4523}} & \textcolor{ImpUp}{\textbf{75.3250}} & \textcolor{ImpUp}{\textbf{30.6065}} \\
 & \quad + scene + caption style GT & \textcolor{ImpUp}{24.2462} & \textcolor{ImpUp}{0.7393} & \textcolor{ImpUp}{0.2022} & \textcolor{ImpUp}{0.1593} & \textcolor{ImpDown}{0.5011} & \textcolor{ImpDown}{71.5270} & \textcolor{ImpDown}{0.5770} & \textcolor{ImpUp}{61.2830} & \textcolor{ImpUp}{74.3444} & \textcolor{ImpUp}{28.9704} \\
\cmidrule(l){2-12}
 & DiT4SR & 22.6539 & 0.6933 & 0.2477 & 0.1830 & 0.4973 & \textbf{70.8502} & 0.6074 & 59.1555 & 73.1471 & 24.5416 \\
 & \quad + caption style GT & \textcolor{ImpUp}{\textbf{23.0546}} & \textcolor{ImpUp}{\textbf{0.7182}} & \textcolor{ImpUp}{\textbf{0.2317}} & \textcolor{ImpUp}{\textbf{0.1797}} & \textcolor{ImpDown}{0.4944} & \textcolor{ImpDown}{70.8355} & \textcolor{ImpDown}{0.5932} & \textcolor{ImpUp}{74.1048} & \textcolor{ImpUp}{\textbf{85.1767}} & \textcolor{ImpUp}{46.7137} \\
 & \quad + scene + caption style GT & \textcolor{ImpUp}{22.6967} & \textcolor{ImpUp}{0.6936} & \textcolor{ImpDown}{0.2481} & \textcolor{ImpDown}{0.1835} & \textcolor{ImpUp}{\textbf{0.5002}} & \textcolor{ImpDown}{70.7191} & \textcolor{ImpUp}{\textbf{0.6103}} & \textcolor{ImpUp}{\textbf{74.1941}} & \textcolor{ImpUp}{84.9031} & \textcolor{ImpUp}{\textbf{47.7010}} \\
\midrule
\multirow{3}{*}{C} & OSEDiff & \textbf{23.9596} & 0.7533 & \textbf{0.2147} & \textbf{0.1666} & 0.4803 & 71.1559 & 0.5672 & 44.6204 & 58.4379 & 10.4372 \\
 & \quad + caption style GT & \textcolor{ImpDown}{23.9394} & \textcolor{ImpUp}{\textbf{0.7546}} & \textcolor{ImpDown}{0.2179} & \textcolor{ImpDown}{0.1695} & \textcolor{ImpUp}{0.4811} & \textcolor{ImpDown}{71.1110} & \textcolor{ImpUp}{0.5676} & \textcolor{ImpUp}{\textbf{51.4657}} & \textcolor{ImpUp}{\textbf{66.4082}} & \textcolor{ImpUp}{\textbf{14.4993}} \\
 & \quad + scene + caption style GT & \textcolor{ImpDown}{23.9081} & \textcolor{ImpDown}{0.7512} & \textcolor{ImpDown}{0.2167} & \textcolor{ImpDown}{0.1677} & \textcolor{ImpUp}{\textbf{0.4872}} & \textcolor{ImpUp}{\textbf{71.1714}} & \textcolor{ImpUp}{\textbf{0.5722}} & \textcolor{ImpUp}{49.3707} & \textcolor{ImpUp}{63.7856} & \textcolor{ImpUp}{13.7094} \\
\midrule
\multirow{6}{*}{E} & TeReDiff & \textbf{23.5188} & \textbf{0.7233} & \textbf{0.2362} & 0.1796 & 0.5638 & 70.5181 & 0.6183 & 51.2383 & 65.1083 & 16.9252 \\
 & \quad + caption style GT & \textcolor{ImpDown}{23.1785} & \textcolor{ImpDown}{0.7176} & \textcolor{ImpDown}{0.2388} & \textcolor{ImpDown}{0.1812} & \textcolor{ImpUp}{0.5694} & \textcolor{ImpUp}{\textbf{71.0449}} & \textcolor{ImpUp}{\textbf{0.6251}} & \textcolor{ImpUp}{\textbf{59.5128}} & \textcolor{ImpUp}{\textbf{72.7708}} & \textcolor{ImpUp}{\textbf{26.7137}} \\
 & \quad + scene + caption style GT & \textcolor{ImpDown}{23.3463} & \textcolor{ImpDown}{0.7209} & \textcolor{ImpDown}{0.2369} & \textcolor{ImpUp}{\textbf{0.1792}} & \textcolor{ImpUp}{\textbf{0.5733}} & \textcolor{ImpUp}{70.8363} & \textcolor{ImpUp}{0.6207} & \textcolor{ImpUp}{57.3934} & \textcolor{ImpUp}{70.6842} & \textcolor{ImpUp}{24.5134} \\
\cmidrule(l){2-12}
 & UniT & 22.7672 & 0.6503 & 0.2814 & 0.2072 & \textbf{0.5159} & \textbf{68.7382} & \textbf{0.6315} & 61.3886 & 76.4538 & 24.1185 \\
 & \quad + caption style GT & \textcolor{ImpUp}{23.1021} & \textcolor{ImpUp}{0.6725} & \textcolor{ImpUp}{0.2694} & \textcolor{ImpUp}{0.1994} & \textcolor{ImpDown}{0.5024} & \textcolor{ImpDown}{68.3612} & \textcolor{ImpDown}{0.6169} & \textcolor{ImpUp}{70.6618} & \textcolor{ImpUp}{82.8848} & \textcolor{ImpUp}{40.4231} \\
 & \quad + scene + caption style GT & \textcolor{ImpUp}{\textbf{23.1256}} & \textcolor{ImpUp}{\textbf{0.6742}} & \textcolor{ImpUp}{\textbf{0.2662}} & \textcolor{ImpUp}{\textbf{0.1975}} & \textcolor{ImpDown}{0.5044} & \textcolor{ImpDown}{68.4496} & \textcolor{ImpDown}{0.6193} & \textcolor{ImpUp}{\textbf{71.1815}} & \textcolor{ImpUp}{\textbf{83.0787}} & \textcolor{ImpUp}{\textbf{41.7489}} \\
\bottomrule
\end{tabular}%
}
\end{table}

\paragraph{Scene description ablation.}
While GT transcripts improve GTTCA, they lower NR-IQA scores for most models (Table~\ref{tab:tab_gt}). We hypothesize that this drop arises because a transcript-only caption departs from the captions these models are trained with, which mostly describe the scene. To test this, we prepend a one-sentence scene description of each LR image, generated by Gemma 4, to the caption-style GT. Table~\ref{tab:supp_tab_rtc_ablation} shows that the scene description largely recovers NR-IQA. The scene description also changes GTTCA. It lowers overall GTTCA by $1.2$ to $2.1$ points for SUPIR, FaithDiff, OSEDiff, and TeReDiff, whereas it leaves DiT4SR nearly unchanged and improves UniT on both levels. We attribute this difference to the text encoders of the underlying text-to-image models. SUPIR, FaithDiff, OSEDiff, and TeReDiff build on models that use only CLIP text encoders, whose short context can truncate or dilute the transcripts that follow the scene description. DiT4SR and UniT build on the same backbone~\cite{sd35}, which additionally uses a T5 text encoder that handles long captions. RTC includes the scene description despite this small cost in GTTCA, since it preserves perceptual quality. A text-to-image backbone that handles long captions may therefore be a good starting point for future TAISR.

\subsection{Observation 2: MLLMs Can Approximate Ground-Truth Transcripts}
\label{sec:supp_rtc_reasoning}

\begin{center}
\small
\setlength{\tabcolsep}{8pt}
\begin{tabular}{l|ccc}
\toprule
Reason type & Image Context & Logical Patterns & Prior Knowledge \\
\midrule
Text Recall & 81.9 & 69.5 & 69.1 \\
\bottomrule
\end{tabular}
\end{center}

We further break down Gemma 4's Level~2 text recall by the annotated cue ($288$ image-context, $246$ logical-pattern, and $414$ prior-knowledge instances, where an instance can carry more than one cue). Recall is highest for image context ($81.9\%$), where the cue is visible elsewhere in the image, and similar for logical patterns ($69.5\%$) and prior knowledge ($69.1\%$). If data contamination drove the performance, prior knowledge, the cue most tied to pre-training data, would likely be read best. Instead, it is read no better than logical patterns. Together with the analysis in Section~\ref{sec:reasoning}, this suggests that Gemma 4's Level~2 recall comes from reasoning over the image rather than from memorization.

\subsection{Method: Reasoning Transfer via Captioning}
\label{sec:supp_rtc_method}

\begin{figure*}[htbp]
    \centering
    \includegraphics[width=\linewidth]{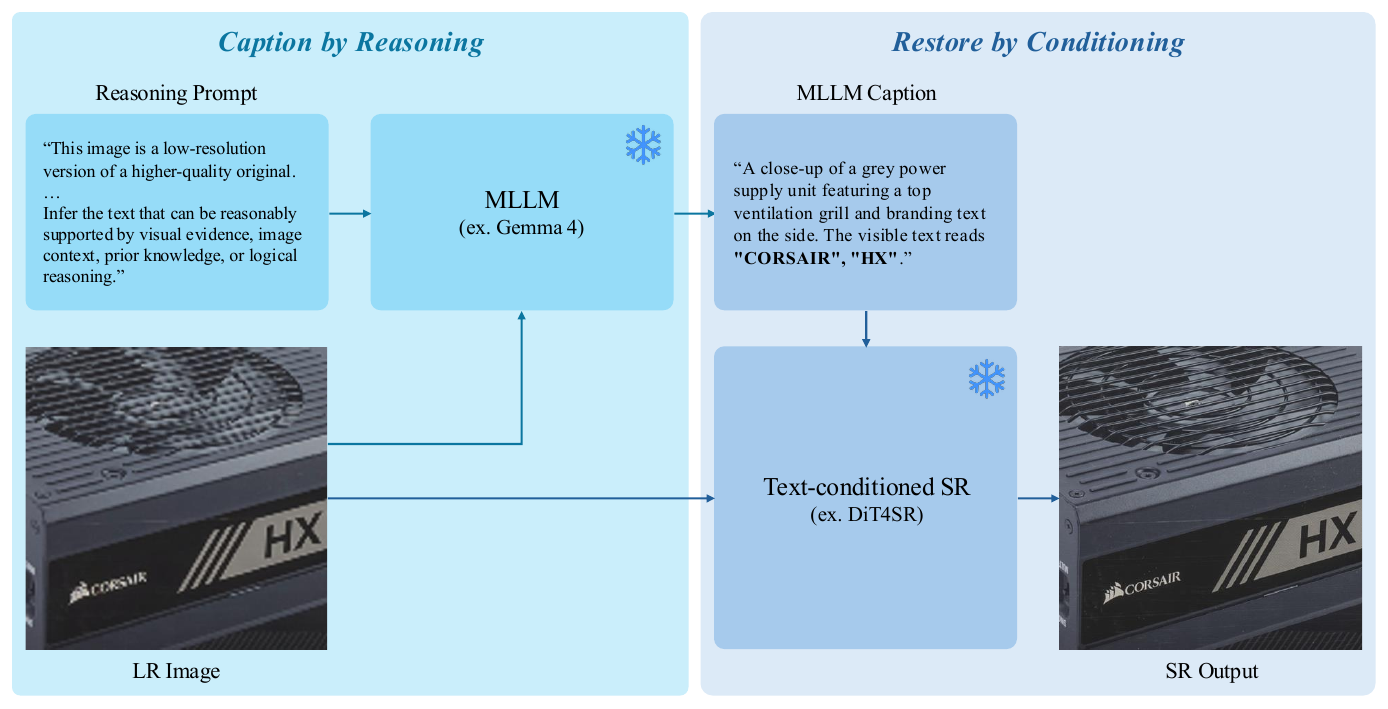}
    \caption{\textbf{Overview of Reasoning Transfer via Captioning (RTC).} (1) Caption by reasoning: a frozen MLLM reads the LR image under the reasoning prompt and returns a two-sentence caption with a scene description and the inferred transcripts. (2) Restore by conditioning: a frozen text-conditioned SR model restores the LR image with this caption in place of its default caption.}
    \label{supp_fig:supp_fig_rtc}
\end{figure*}

The reasoning prompt asks the MLLM for a two-sentence caption. The first sentence describes the scene, which avoids most of the NR-IQA drop observed with transcript-only captions (Appendix~\ref{sec:supp_rtc_injection}). The second sentence lists the transcripts inferred from the LR image. Figure~\ref{supp_fig:supp_fig_rtc} illustrates the overall pipeline with an example caption.

\section{Additional Experiments}
\label{sec:supp_exp}

This section shows that RTC generalizes to existing TAISR benchmarks, is robust to imperfect captions, and remains effective on Level~3 text.

\begin{table}[t]
\centering
\setlength{\aboverulesep}{0pt}\setlength{\belowrulesep}{0pt}\renewcommand{\arraystretch}{1.1}
\caption{\textbf{RTC improves text restoration on existing TAISR benchmarks.} Text-spotting metric (TESTR~\cite{testr}) measured on RealText and the three degradation levels of SA-Text-test, comparing each text-conditioned SR model with and without RTC. The levels of SA-Text-test denote degradation strengths and are unrelated to the difficulty levels of ReasonText. \textcolor{ImpUp}{Blue} marks an improvement over the baseline. The HR oracle row reports the score on the HR images for reference.}
\label{tab:supp_tab_exp_gen_overall}
\begin{tabular}{cl|c|ccc}
\toprule
\multirow{2}{*}{Group} & \multirow{2}{*}{Model} & \multirow{2}{*}{RealText} & \multicolumn{3}{c}{SA-Text-test} \\
\cmidrule(lr){4-6}
& & & Level~1 & Level~2 & Level~3 \\
\midrule
\rowcolor{cyan!20}
\multicolumn{2}{c|}{HR (oracle)} & 75.4443 & 76.2919 & 76.2919 & 76.2919 \\
\midrule
\multirow{6}{*}{B} & SUPIR & 41.6263 & 23.5721 & 20.2629 & 12.2847 \\
 & \quad + RTC & \textcolor{ImpUp}{50.9962} & \textcolor{ImpUp}{34.0435} & \textcolor{ImpUp}{28.4225} & \textcolor{ImpUp}{18.5857} \\
\cmidrule(l){2-6}
 & FaithDiff & 43.4572 & 23.5721 & 20.7162 & 13.3726 \\
 & \quad + RTC & \textcolor{ImpUp}{49.9731} & \textcolor{ImpUp}{28.4678} & \textcolor{ImpUp}{26.5186} & \textcolor{ImpUp}{18.7670} \\
\cmidrule(l){2-6}
 & DiT4SR & 46.9575 & 25.1133 & 21.9855 & 15.2765 \\
 & \quad + RTC & \textcolor{ImpUp}{56.4890} & \textcolor{ImpUp}{36.4461} & \textcolor{ImpUp}{31.6410} & \textcolor{ImpUp}{21.5775} \\
\midrule
\multirow{2}{*}{C} & OSEDiff & 41.4647 & 19.6283 & 16.9991 & 10.9248 \\
 & \quad + RTC & \textcolor{ImpUp}{46.0420} & \textcolor{ImpUp}{24.0254} & \textcolor{ImpUp}{20.8069} & \textcolor{ImpUp}{15.1405} \\
\midrule
\multirow{4}{*}{E} & TeReDiff & 46.2036 & 23.3908 & 21.5775 & 15.5938 \\
 & \quad + RTC & \textcolor{ImpUp}{49.4346} & \textcolor{ImpUp}{26.9719} & \textcolor{ImpUp}{24.4787} & \textcolor{ImpUp}{17.2711} \\
\cmidrule(l){2-6}
 & UniT & 47.0113 & 26.2919 & 23.5267 & 16.7724 \\
 & \quad + RTC & \textcolor{ImpUp}{54.8196} & \textcolor{ImpUp}{34.0888} & \textcolor{ImpUp}{30.3717} & \textcolor{ImpUp}{20.2629} \\
\bottomrule
\end{tabular}%
\end{table}

\paragraph{RTC also improves text restoration on existing TAISR benchmarks.}
We further evaluate RTC on the two existing TAISR benchmarks, SA-Text-test and RealText. SA-Text-test consists of $1{,}000$ LR--HR pairs at three synthetic degradation strengths, denoted Levels 1-3 in Table~\ref{tab:supp_tab_exp_gen_overall}, where a higher level means stronger degradation. These levels are unrelated to the difficulty levels of ReasonText. RealText consists of $847$ LR--HR pairs of scene images with real-world degradations. Since both benchmarks are annotated by fully automatic pipelines without human verification, their labels contain non-trivial noise (Section~\ref{subsec:benchmark}). This noise would directly affect GTTCA, which relies on the GT polygons and transcripts. We therefore evaluate each model with and without RTC under the original text spotting protocol of these benchmarks.

Table~\ref{tab:supp_tab_exp_gen_overall} reports the results on RealText and on the three degradation levels of SA-Text-test. In all four settings, RTC improves all six text-conditioned SR models without exception, indicating that its gains are not an artifact of a particular benchmark or metric. Moreover, UniT outperforms DiT4SR in every setting without RTC, whereas DiT4SR\,+\,RTC outperforms UniT\,+\,RTC in every setting. This reproduces the reversal observed on ReasonText (Section~\ref{sec:main_results}) and again suggests that reasoning should be a core axis of text-aware design.

\paragraph{RTC is robust to imperfect captions.}
Since RTC feeds the SR model a caption predicted by an MLLM, one might worry that a wrong prediction would degrade restoration. To examine this, we split the $709$ Level~2 instances by whether the RTC caption contains the correct GT transcript, into a Gemma correct subset ($499$ instances) and a Gemma wrong subset ($210$ instances), and measure Level~2 GTTCA on each subset before and after RTC (Table~\ref{tab:supp_tab_exp_robust}(a)). Even on the Gemma wrong subset, RTC improves GTTCA for all six SR models. Half of the wrong captions are near-misses that differ from the GT transcript by at most two letters, and the gain concentrates on them ($17.8\%$ to $25.9\%$), since the SR models can still restore the exact text from a near-correct caption together with the LR image (Table~\ref{tab:supp_tab_exp_robust}(b)). For captions farther from the GT transcript, RTC still does not lower GTTCA on average ($10.2\%$ to $11.6\%$). RTC is therefore robust to imperfect captions, which matters in practice since the MLLM may misread degraded text.

\begin{table}[t]
\centering
\setlength{\aboverulesep}{0pt}\setlength{\belowrulesep}{0pt}\renewcommand{\arraystretch}{1.2}
\caption{\textbf{RTC is robust to caption errors.} (a) Level~2 GTTCA before (base) and after RTC (+RTC), split by whether the RTC caption contains the correct GT transcript. RTC improves GTTCA for all six SR models even when Gemma 4 is wrong. (b) The Gemma wrong subset pooled over the six SR models and split by how far the caption is from the GT transcript. The gain concentrates on near-miss captions that differ by at most two letters, from which the SR models can still restore the exact text. \textcolor{ImpUp}{Blue} marks an improvement over the baseline.}
\label{tab:supp_tab_exp_robust}
\begin{subtable}[t]{0.55\textwidth}
\centering
\caption{Per-model, split by caption correctness}
\small
\setlength{\tabcolsep}{5pt}
\begin{tabular}{cl|cc|cc}
\toprule
\multirow{3}{*}{Group} & \multirow{3}{*}{Model} & \multicolumn{2}{c|}{Gemma correct} & \multicolumn{2}{c}{Gemma wrong} \\
& & \multicolumn{2}{c|}{($n{=}499$)} & \multicolumn{2}{c}{($n{=}210$)} \\
\cmidrule(lr){3-4}\cmidrule(lr){5-6}
& & \multicolumn{1}{c}{base} & \multicolumn{1}{c|}{+RTC} & \multicolumn{1}{c}{base} & \multicolumn{1}{c}{+RTC} \\
\midrule
\multirow{3}{*}{B} & SUPIR & 25.7 & \textcolor{ImpUp}{46.5} & 15.2 & \textcolor{ImpUp}{20.5} \\
 & FaithDiff & 21.2 & \textcolor{ImpUp}{33.1} & 15.2 & \textcolor{ImpUp}{19.0} \\
 & DiT4SR & 26.9 & \textcolor{ImpUp}{55.3} & 19.0 & \textcolor{ImpUp}{23.3} \\
\midrule
C & OSEDiff & 11.8 & \textcolor{ImpUp}{13.4} & 7.1 & \textcolor{ImpUp}{11.0} \\
\midrule
\multirow{2}{*}{E} & TeReDiff & 18.8 & \textcolor{ImpUp}{25.9} & 12.4 & \textcolor{ImpUp}{17.1} \\
 & UniT & 28.1 & \textcolor{ImpUp}{51.1} & 14.8 & \textcolor{ImpUp}{21.4} \\
\bottomrule
\end{tabular}
\label{tab:supp_tab_exp_robust_a}
\end{subtable}
\hfill
\begin{subtable}[t]{0.43\textwidth}
\centering
\caption{Gemma wrong subset, by caption distance}
\small
\setlength{\tabcolsep}{5pt}
\begin{tabular}{l|ccc}
\toprule
Wrong L2 caption & base & +RTC & $\Delta$ \\
\midrule
near-miss ($\leq$2 letters) & 17.8 & \textcolor{ImpUp}{25.9} & \textcolor{ImpUp}{+8.1} \\
far ($>$2 letters) & 10.2 & \textcolor{ImpUp}{11.6} & \textcolor{ImpUp}{+1.4} \\
\bottomrule
\end{tabular}
\label{tab:supp_tab_exp_robust_b}
\end{subtable}
\end{table}

\begin{table}[t]
\centering
\setlength{\aboverulesep}{0pt}\setlength{\belowrulesep}{0pt}\renewcommand{\arraystretch}{1.2}
\caption{\textbf{RTC is effective at every difficulty level, including Level~3.} (a) GTTCA before (base) and after RTC (+RTC) on each difficulty level. RTC improves all six SR models on Level~1, Level~2, and even Level~3, though the Level~3 gains are smaller on average. Mean $\Delta$ is the average gain over the six models. (b) Text recall of Gemma 4 with the reasoning prompt on each level. \textcolor{ImpUp}{Blue} marks an improvement over the baseline.}
\label{tab:supp_tab_exp_level3}
\begin{subtable}[t]{0.66\textwidth}
\centering
\caption{GTTCA per backbone and difficulty level}
\small
\setlength{\tabcolsep}{5pt}
\begin{tabular}{cl|cc|cc|cc}
\toprule
\multirow{2}{*}{Group} & \multirow{2}{*}{Model} & \multicolumn{2}{c|}{Level~1} & \multicolumn{2}{c|}{Level~2} & \multicolumn{2}{c}{Level~3} \\
\cmidrule(lr){3-4}\cmidrule(lr){5-6}\cmidrule(lr){7-8}
& & \multicolumn{1}{c}{base} & \multicolumn{1}{c|}{+RTC} & \multicolumn{1}{c}{base} & \multicolumn{1}{c|}{+RTC} & \multicolumn{1}{c}{base} & \multicolumn{1}{c}{+RTC} \\
\midrule
\multirow{3}{*}{B} & SUPIR & 68.5 & \textcolor{ImpUp}{78.8} & 22.6 & \textcolor{ImpUp}{38.8} & 4.9 & \textcolor{ImpUp}{11.8} \\
 & FaithDiff & 65.2 & \textcolor{ImpUp}{74.0} & 19.5 & \textcolor{ImpUp}{28.9} & 2.7 & \textcolor{ImpUp}{5.8} \\
 & DiT4SR & 73.1 & \textcolor{ImpUp}{84.8} & 24.5 & \textcolor{ImpUp}{45.8} & 6.2 & \textcolor{ImpUp}{18.3} \\
\midrule
C & OSEDiff & 58.4 & \textcolor{ImpUp}{64.3} & 10.4 & \textcolor{ImpUp}{12.7} & 1.4 & \textcolor{ImpUp}{2.0} \\
\midrule
\multirow{2}{*}{E} & TeReDiff & 65.1 & \textcolor{ImpUp}{70.1} & 16.9 & \textcolor{ImpUp}{23.3} & 2.6 & \textcolor{ImpUp}{5.0} \\
 & UniT & 76.5 & \textcolor{ImpUp}{82.6} & 24.1 & \textcolor{ImpUp}{42.3} & 5.1 & \textcolor{ImpUp}{14.1} \\
\midrule
\multicolumn{2}{c|}{Mean $\Delta$} & \multicolumn{2}{c|}{\textcolor{ImpUp}{$8.0$}} & \multicolumn{2}{c|}{\textcolor{ImpUp}{$12.3$}} & \multicolumn{2}{c}{\textcolor{ImpUp}{$5.7$}} 
\\
\bottomrule
\end{tabular}
\label{tab:supp_tab_exp_level3_a}
\end{subtable}
\hfill
\begin{subtable}[t]{0.32\textwidth}
\centering
\caption{Gemma~4 text recall by level}
\small
\setlength{\tabcolsep}{6pt}
\begin{tabular}{ccc}
\toprule
Level~1 & Level~2 & Level~3 \\
\midrule
86.7 & 70.4 & 40.2 \\
\bottomrule
\end{tabular}
\label{tab:supp_tab_exp_level3_b}
\end{subtable}
\end{table}

\paragraph{RTC remains effective on Level 3 text.}
ReasonText's Level~3 comprises text that none of our annotators could read even from the full LR patch, so it is excluded from our main evaluation. Nonetheless, since RTC injects an MLLM caption that may be wrong, a natural concern is that it might harm this hardest case. Table~\ref{tab:supp_tab_exp_level3} (a) shows that it does not. RTC improves GTTCA on Level~3 for all six SR models, though by less than on the other levels on average. These gains are possible because Gemma 4 with the reasoning prompt still recalls $40.2\%$ of Level~3 text (Table~\ref{tab:supp_tab_exp_level3}(b)). We attribute this to how Level~3 is defined. Level~3 means that none of our annotators could read the text, not that the text is intrinsically unrecoverable. This set is therefore heterogeneous. Some instances, such as serial numbers, are genuinely unrecoverable, while others might have been read by a larger annotator pool or by domain experts with the relevant prior knowledge. Gemma 4 likely reads some instances of the latter kind, reflecting the limitation of our annotation protocol noted in Section~\ref{sec:conclusion}.

\section{Limitations and Broader Impact}
\label{sec:supp_limitations}

\paragraph{Limitations.}
We summarize the limitations discussed in Section~\ref{sec:conclusion}. First, ReasonText is human-curated and built on existing real-world SR benchmarks, so it is limited in size and scope, with $513$ LR--HR pairs of natural images and only alphanumeric text annotated. Second, even the best model with RTC remains far below the HR oracle on Level~2 ($45.84\%$ versus $90.69\%$). Since the same model reaches at most $47.70\%$ even with GT transcripts (Table~\ref{tab:supp_tab_rtc_ablation}), most of this gap lies not in reading the text but in rendering it, and a stronger MLLM alone would bring limited gains. Third, RTC applies only to text-conditioned SR models and cannot be used with SR models that have no text channel.

\paragraph{Broader Impact.}
By enabling SR models to recover text that was previously unreadable from local pixels alone, this work has the potential to benefit a wide range of socially valuable applications. Improved restoration of degraded text can (i) support accessibility for users with low vision by making blurred signs, labels, and documents legible, (ii) aid cultural and historical preservation by helping recover text in old photographs, archival scans, and damaged manuscripts, and (iii) contribute to public safety and navigation through clearer recovery of road signs, transit information, and emergency notices captured under poor imaging conditions. More broadly, our findings suggest that reasoning, rather than purely local perception, is a productive direction for restoration tasks, encouraging the community to explore restoration pipelines that integrate world knowledge and contextual understanding. At the same time, text restored through reasoning is a plausible inference rather than evidence, and it may occasionally diverge from the true content. Such outputs should not be treated as ground truth in high-stakes settings such as forensic analysis or document verification, and the ability to read degraded text calls for care with images containing private information, such as license plates or personal documents. The decoupled design of RTC helps mitigate these risks, since the MLLM caption exposes the inferred text and makes it possible to trace which component caused an error. Overall, we hope ReasonText, GTTCA, and RTC together provide a constructive foundation for building TAISR systems that are not only sharper, but also more thoughtful about the text they restore.

\clearpage
\newpage

\end{document}